\pdfoutput=1

\documentclass[nohyperref]{article}

\usepackage{microtype}
\usepackage{graphicx}
\usepackage{booktabs} 

\usepackage{hyperref}

\usepackage[accepted]{icml2022}

\usepackage{times}
\usepackage{latexsym}
\usepackage[utf8]{inputenc} 
\usepackage[T1]{fontenc}    
\usepackage{url}            
\usepackage{booktabs}       
\usepackage{amsfonts}       
\usepackage{nicefrac}       
\usepackage{microtype}      
\usepackage{graphicx}
\usepackage{amsmath}
\usepackage{amssymb}
\usepackage{cleveref}       

\usepackage{xspace}
\usepackage{bm}
\usepackage{tabu}
\usepackage{xcolor}
\usepackage{caption}
\usepackage{subcaption}
\usepackage{multirow}
\usepackage{appendix}
\usepackage{etoolbox}
\usepackage{float}
\usepackage{dsfont}
\usepackage{nicefrac}

\usepackage{lipsum}

\usepackage{listings} 
\usepackage{svg}
\usepackage{setspace}
\usepackage{tabularx}
\usepackage{tabulary}

\BeforeBeginEnvironment{appendices}{\clearpage}

 \hypersetup{
     colorlinks=false,
     linkcolor=blue,
     filecolor=blue,
     citecolor = black,      
     urlcolor=cyan,
     }

\makeatletter
\newcommand\footnoteref[1]{\protected@xdef\@thefnmark{\ref{#1}}\@footnotemark}
\makeatother

\newcommand{\ra}[1]{\renewcommand{\arraystretch}{#1}}

\newcommand{\blap}[1]{\smash[b]{\begin{tabular}[t]{@{}l@{}}#1\end{tabular}}}

\newcommand{\ie}{\textit{i.e.}}
\newcommand{\eg}{\textit{e.g.}}

\newcommand{\tabref}{Table~}
\newcommand{\appref}{Appendix~}
\newcommand{\secref}{Sec.~}
\newcommand{\eq}{Eq.~}

\newcommand{\TM}{TLMs\xspace}
\newcommand{\oneTM}{TLM\xspace}
\newcommand{\ap}{\text{AP}}

\newcommand{\maxap}{\max_m\{\ap_m^c\}}

\newcommand{\bx}{\bm{x}}
\newcommand{\bz}{\bm{z}}

\newcommand{\by}{\bm{b}}

\newcommand{\np}{N^+_c}
\newcommand{\nm}{N^-_c}

\newcommand{\wordnet}[1]{http://wordnetweb.princeton.edu/perl/webwn?s=#1\&sub=Search+WordNet\&o2=\&o0=1\&o8=1\&o1=1\&o7=\&o5=\&o9=\&o6=1\&o3=\&o4=\&h=0000000}

\newcommand{\gen}[2]{{\fontsize{#1}{7}\textsf{#2}\selectfont}}

\newcommand{\pplm}{PPLM-BoW\xspace}
\newcommand{\fudge}{FUDGE\xspace}

\definecolor{codegreen}{rgb}{0,0.6,0}
\definecolor{codegray}{rgb}{0.5,0.5,0.5}
\definecolor{codepurple}{rgb}{0.58,0,0.82}
\definecolor{backcolour}{rgb}{0.95,0.95,0.92}

\lstdefinestyle{mystyle}{
  backgroundcolor=\color{backcolour}, commentstyle=\color{codegreen},
  keywordstyle=\color{magenta},
  numberstyle=\tiny\color{codegray},
  stringstyle=\color{codepurple},
  basicstyle=\ttfamily\scriptsize,
  breakatwhitespace=false,         
  breaklines=true,                 
  captionpos=b,                    
  keepspaces=true,                 
  numbers=left,                    
  numbersep=5pt,                  
  showspaces=false,                
  showstringspaces=false,
  showtabs=false,                  
  tabsize=1
}

\icmltitlerunning{Self-conditioning Pre-Trained Language Models}

\begin{document}

\twocolumn[
\icmltitle{Self-conditioning Pre-Trained Language Models}



\icmlsetsymbol{equal}{*}

\begin{icmlauthorlist}
\icmlauthor{Xavier Suau}{Apple}
\icmlauthor{Luca Zappella}{Apple}
\icmlauthor{Nicholas Apostoloff}{Apple}
\end{icmlauthorlist}

\icmlaffiliation{Apple}{Apple}

\icmlcorrespondingauthor{Xavier Suau}{xsuaucuadros@apple.com}

\icmlkeywords{Generative modelling, language model, bias, fairness, controllability}

\vskip 0.3in
]



\printAffiliationsAndNotice{}  

\begin{abstract}

In this paper we aim to investigate the mechanisms that guide text generation with pre-trained Transformer-based Language Models (\TM). Grounded on the Product of Experts formulation by \citet{Hinton:ICANN:1999}, we describe a generative mechanism that exploits expert units which naturally exist in \TM. Such units are responsible  for detecting concepts in the input and conditioning text generation on such concepts. We describe how to identify expert units and how to activate them during inference in order to induce any desired concept in the generated output. We find that the activation of a surprisingly small amount of units is sufficient to steer text generation (as little as 3 units in a model with 345M parameters). While the objective of this work is to learn more about how \TM work, we show that our method is effective for conditioning without fine-tuning or using extra parameters, even on fine-grained homograph concepts. Additionally, we show that our method can be used to correct gender bias present in the output of \TM and achieves gender parity for all evaluated contexts. We compare our method with \fudge \cite{Yang2021} and \pplm \cite{Dathathri2020Plug}, and show that our approach is able to achieve gender parity at a lower perplexity. The proposed method is accessible to a wide audience thanks to its simplicity and minimal compute needs. The findings in this paper are a step forward in understanding the generative mechanisms of \TM.

\end{abstract}



\section{Introduction }
\label{sec:introduction}

Natural Language Processing (NLP) is evolving at a fast pace. For instance, language models \cite{Bengio:JMLR:2003} based on the Transformer \cite{Vaswani:NIPS:2017} architecture (\TM) achieve impressive performance in many tasks, including text generation \cite{Radford:ARXIV:2019,brown2020language}. Despite their success, the mechanisms that govern and control text generation using \TM remain unclear. Additionally, \TM have two important drawbacks: (1) conditioning these models to constrain the content of their generation requires either expensive re-training \cite{Keskar:Arxiv:2019} or the use of additional parameters \cite{Dathathri2020Plug,zhang2020side,Yang2021}; (2) \TM might inherit and perpetuate biases present in the training data, which can have a negative social impact \cite{sheng2019woman,AbubakarNature2021}, especially when \TM are deployed in commercial applications.

This paper investigates the internal mechanisms used by \TM to control text generation. Theoretically grounded  on the Product of Experts formulation by \citet{Hinton:ICANN:1999}, we describe a generative mechanism based on the presence of \emph{expert units} (neurons) in pre-trained \TM. In this mechanism, expert units capture the presence of a concept in the \oneTM input. At generation time, the \oneTM relies on the activation of such expert units to produce text with a specific concept. In order to exploit the proposed generative mechanism, we first identify expert units based on their ability to detect a concept in the input with a given average precision. Our approach finds expert units in a scalable manner for a variety of concepts. Then, we apply a post-hoc intervention upon those units which increases the presence of a concept in the generated text, independently of the concepts present in the input. We show that intervening on as little as 3 units can be sufficient to steer the generation towards a desired concept. Our method does not require fine-tuning or using additional parameters\footnote{Code available at \href{https://github.com/apple/ml-selfcond}{https://github.com/apple/ml-selfcond}}.

\secref\ref{sec:soa} contains a literature review of relevant works. In \secref\ref{sec:sentenceconcepts} we define the term \textit{concept} and we describe its formal representation. In \secref\ref{sec:method} we explain our algorithm to exploit the proposed generative mechanism by finding and activating expert units for a specific concept.
In \secref\ref{sec:results_generation} we apply our method to condition \TM on a variety of concepts (including fine-grained homographs) and we provide qualitative text generation results.
In \secref\ref{sec:results_parity} we analyze the performance of our method for gender bias mitigation. We compare our method with the state of the art methods \fudge \cite{Yang2021} and PPLM \cite{Dathathri2020Plug} in its Bag-of-Words version (\pplm). \pplm and our method are the only works that achieve conditional generation without additional parameters. Our method outperforms the compared algorithms, achieving \emph{generative parity} (\ie,  the \oneTM generates sentences with equal probability of containing specific concepts) with higher quality, measured by a lower perplexity and less same-word repetition. 
Interestingly, we achieve parity by intervening on very few expert units (a median of 15 units, representing 0.0067\% of the model units analyzed). 
We further show that the conditioning strength required to achieve parity with our method is correlated with the intrinsic bias of the model, which facilitates the choice of the conditioning parameters. Lastly, we validate empirically our choice of expert units. In \secref\ref{sec:limitations} we discuss the limitations and potential improvements of our work. Finally, conclusions are drawn in \secref\ref{sec:conclusions}.

\section{Related work
}
\label{sec:soa}

\paragraph{Expert units.} The use of expert units has been previously explored in the image domain \cite{Bau:CVPR:2017,Bau:ICLR:2019,Fong:CVPR:2018}.
Our work is inspired by this body of research. However, adapting it to the NLP domain has required redefining what an expert unit is, how to find it, and how to control it. \citet{radford2017learning} finds an expert unit for sentiment (the \textit{sentiment neuron}) in LSTM \cite{hochreiter1997long} representations. It does so via L1 regularization of a logistic regression classifier on top of the representations. Our work is not limited to sentiment, and it can scale to much larger models such as \TM.
\vspace{-2mm}
\paragraph{Product of Experts.} Some recent works propose conditioning strategies with minimal intervention on the \oneTM. For instance, PPLM \cite{Dathathri2020Plug} exploits the Product of Experts (PoE) formulation \cite{Hinton:ICANN:1999} and does not require re-training. They steer the latent variables during generation to maximize both a conditional expert (modelled with an external network) and the unconditional expert. The steering is performed using the gradients from the external network. In the \pplm form, the conditional expert is a Bag-of-Words (BoW) model, which does not require any training parameter. Side tuning \cite{zhang2020side} adds a side model that learns a residual on top of the original model. Similarly, \cite{zeldes2020technical} supplements the pre-trained \oneTM with an external model that shifts the output distribution. Recently, FUDGE \cite{Yang2021} adjusts the output probabilities of a \oneTM by training a discriminator model that predicts whether a topic will appear in the future. In FUDGE, the discriminator can also be trained to condition formality or poetry.  All these methods follow the PoE framework (explicitly, or implicitly). Our formulation also adopts the PoE framework, with a key difference: we consider that the conditional PoE expert already exists in the \oneTM rather than using external models. We propose a way to identify the PoE conditional expert that does not involve computing gradients or using additional parameters. This makes our solution simple and accessible to a wider audience,  and also unveils aspects of how generative mechanisms of \TM work.
\vspace{-2mm}
\paragraph{Conditioned text generation.} Most methods tackling conditioned text generation are based on training dedicated architectures. 
In \cite{Chen:NAACL:2019}, two latent embeddings representing syntax and semantics are inferred enforcing disentanglement. This allows conditioning on an arbitrary combination of syntax and semantics. Similarly, \cite{Romanov:NAACL:2019} disentangle meaning and form with an adversarial training approach. The work in \cite{Hu:ICML:2017} combines a Variational Auto-Encoder \cite{kingma2013auto} with discriminators of specific attributes, and shows results controlling sentiment and tense. In \cite{peng-etal-2018-towards}, human specified control factors are extracted from data by an analyzer model. Such factors are used at generation time to control the story ending valence (sad or happy endings). In CTRL \cite{Keskar:Arxiv:2019}, training sentences are prepended with a control code, which allows conditioning at test time.  The work in \cite{schiller2020aspect} builds on CTRL allowing the controlled generation of arguments for specific contexts and aspects. 

Although effective, all these methods need the conditioning to be known before the model is trained, require large amounts of data, and suffer from the computational complexities typical of \TM training.
One of the advantages of our approach is that a concept is encoded solely by a set of examples. 
Extending the number of controllable concepts (at any time) is as simple as collecting positive and negative examples for the new concepts.

\section{Concepts as binary sentence datasets 
}
\label{sec:sentenceconcepts}
Throughout this paper we refer to \textit{concept} as an ``abstract idea'' that can be described with a set of examples. We extend \cite{kim2018interpretability} to the NLP domain by describing a concept $c$ with a dataset  $\{\bx_i^c, b_i^c\}_{i=1}^N$ of $N = \np+\nm$ sentences. The $\np$ positive sentences contain $c$ (\ie, $b_i^c=1$), and the $\nm$ negative sentences do \emph{not} contain $c$ (\ie, $b_i^c=0$). Each sentence $\bx_i^c$ is padded to a length $T$.

Such flexible definition allows diverse types of concepts to be represented. They can be broad such as \textit{sport} or more precise one such as \textit{football}, \textit{world cup}, \textit{national football team}, \textit{player}, etc. Our definition also allows representing abstract concepts (\eg, sentiment) or more concrete ones using Bags-of-Words. One interesting aspect of this representation is that we can distinguish homographs, \eg, we can represent the concept \emph{note} ``a reminder'' differently from  \emph{note} ``a tone of certain pitch''.

\setlength{\belowdisplayskip}{3mm} \setlength{\belowdisplayshortskip}{3mm}
\setlength{\abovedisplayskip}{3mm} \setlength{\abovedisplayshortskip}{3mm}

 \section{Method}
 \label{sec:method}
 
\subsection{Generative mechanism based on expert units}
\label{sec:mechanism}
Language models are generative models that can generate text consistent with linguistic rules. More formally, autoregressive language models maximize the probability of a sentence $\bx = \{\bx_i\}$ as $p(\bx) = p(\bx_1,\ldots,\bx_{T}) = \prod_{t=1}^T p(\bx_t | \bx_{<t})$ \cite{Bengio:JMLR:2003}.
A conditional generative model maximizes the joint distribution $p(\bx, y)=p(y|\bx)p(\bx)$, where $\bx$ is the generated sentence and $y$ is a conditional variable (\ie, a concept in $\bx$). \citet{Hinton:ICANN:1999} proposed to interpret this equation as a \textit{Product of Experts}. The same interpretation was adopted in \cite{Dathathri2020Plug,Yang2021} for conditioned text generation, where the conditional model $p(y|\bx)$ determines the condition for generation, while $p(\bx)$ ensures that the generated sequence lies within the manifold of sentence distributions.
In conditioned generation, rather than jointly sampling $\bx$ and $y$, we set the condition $y=c$ beforehand, thus 
\begin{equation}
\label{eq:conditioned_gen}
p(\bx|y=c) \propto p(y=c|\bx)p(\bx).
\end{equation}
Unlike \citet{Dathathri2020Plug} and \citet{Yang2021} that implement the conditional model $p(y=c|\bx)$ with an external network, we hypothesize that \textit{the internal generative mechanism of \TM exploits a conditional model $p(y=c|\bx)$ that already exists \underline{within} the same model}. Therefore, \TM naturally obey a factorised conditional generation like the one in \eq\eqref{eq:conditioned_gen}, and are able to maximize $p(\bx|y=c)$ by exploiting their internal conditional model.  
The quality of the conditional model will dictate the extent to which a concept can be controlled during generation.
On the other hand, a good $p(\bx)$ is also required to ensure that the generated text stays within the language manifold; failing to do so would lead to sentences that maximize  $p(\bx|y=c)$ but are not linguistically correct.

\subsection{Self-conditioning method}
\label{sec:finding}
We denote as \textit{expert units} those neurons that contribute to the conditional model $p(y=c|\bx)$ in \eq\eqref{eq:conditioned_gen}.
Therefore, we propose to identify expert units as neurons whose  response can be used as a predictor for the presence of a concept in the input.
Formally, let $\bz_m^c = \{z^c_{m,i}\}_{i=1}^N$ be the outputs of neuron $m$ to sentences $\{\bx_i^c\}$. We treat $\bz_m^c$ as prediction scores for the task $\by^c = \{b_i^c\}_{i=1}^N$. Thus, we measure the \textit{expertise} of a unit $m$ for the task $\by^c$ with its the Average Precision ($\ap_m^c$), which is area under the precision-recall curve, so that $\ap_m^c = \text{AP}(\bz_m^c, \by^c) \in [0, 1]$.  For each concept $c$ we measure the $\ap_m^c$ for all units and layers and rank them from the highest to the lowest level of expertise. Note that, to be agnostic with respect to the sequence length, the output of each layer is max-pooled  across the temporal dimension (ignoring pad tokens).

To induce the presence of a concept $c$ during text generation, we set the responses of the top experts, irrespective of their input, to their typical values measured when $c$ is present. Borrowing from the causality literature \cite{Pearl:Causality}, we define the intervention on $k$ expert units as a $do(c,k)$ operation on the model responses at inference time. Let $\mathcal{Q}_k$ be the indices of the top-$k$ experts, then the operation in \eq\eqref{eq:forcing} manipulates the responses of the top-$k$ experts by setting them to their expected value for concept $c$:
\begin{equation}
\small
\label{eq:forcing}
    do(c, k): \{\bz_{m}^c := E_{\bx^c}\big[\bz_{m}^c \,|\, \by^c = 1\big]\; \forall m \in \mathcal{Q}_k \}.
\end{equation}
Note that the expectation in \eq\eqref{eq:forcing} can be approximated as $E_{\bx^c}\big[\bz_m^c \,|\, \by^c = 1\big] \approx \sum_{i=1}^{\np} z_{m,i}^c / \np$ (ignoring negative examples). See \appref\ref{app:code} for a PyTorch \cite{pytorch2019} code example that implements  \eq\eqref{eq:forcing}.

In order to maximize \eq\eqref{eq:conditioned_gen} one can maximize $p(y=c|\bx)$ while keeping $p(\bx)$ unchanged. This is the case for pre-trained models, since we cannot improve $p(\bx)$ without re-training or fine-tuning the model. We propose to maximize $p(y=c|\bx)$ by increasing the number of experts $k$ when applying the $do(c,k)$ intervention, \eq~\eqref{eq:forcing}. This allows controlling the ``amount'' of concept in the generated sentences via the parameter $k$.
Although such intervention modifies the model behavior, $p(\bx)$ should be minimally affected, since $k << M$ ($M$ being the total number of units available). Larger values of $k$ will eventually degrade $p(\bx)$ over $p(y=c|\bx)$ and the conditioned generative probability $p(\bx|y=c)$ will collapse.

%

Autoregressive language models generate text via sequential decoding, which ties the presence of a concept in the input and the output of a \oneTM. 
For example, words related to \textit{football} are more likely when the input is about \textit{football}. The $do(c,k)$ operation in \eq~\eqref{eq:forcing} sets the responses of expert units to the values they typically have when the concept is present. Doing so, we artificially simulate the presence of a concept in the input, to effectively induce the model to behave as if the concept is present in the input.


Our method exploits the natural generative mechanism of \TM based on expert units.
This means that we condition the model using its own knowledge (self-conditioning), without the use of any external model or auxiliary training variables, and without fine-tuning.
The results in \secref\ref{sec:generation} confirm our hypothesis that the conditional model exists within the \oneTM, and that the model leverages it to self-condition generation. Specific results in \secref\ref{sec:units_choice} also validate our ranking of expert units.

\section{Experimental analysis}
\label{sec:generation}

We divide our analysis in four sections. First in \secref\ref{sec:results_generation} we show examples of self-conditioned generation. In \secref\ref{sec:results_parity}, we show how our technique can be used to achieve gender parity in TLM's text generation, and we compare it with two state of art techniques for conditioning \TM: \fudge \cite{Yang2021} and \pplm \cite{Dathathri2020Plug}. Differences between these methods and ours are discussed in \secref\ref{sec:differences}. Lastly, in \secref\ref{sec:units_choice} we show that the way we identify and rank expert units is effective to control text generation.
In all our experiments the decoding strategy is by top-$n$ sampling with $n=10$ as \citet{Yang2021}. Details on the layers analyzed in \oneTM architectures are shown in \appref\ref{app:layers_schema}.


We construct our concept dataset leveraging the OneSec dataset \cite{Scarlini:ACL:2019}, which contains sentences with one keyword annotated with a WordNet sense. We chose OneSec because it is composed of Wikipedia articles, a corpus that was not used for the training of the models used in our experiments (GPT2-M and GPT2-L \cite{Radford:ARXIV:2019}). Note that our method is not limited by the choice of a specific data source.

We limit the data per concept to $100 \leq \np \leq 1000$ and $100 \leq \nm \leq 1000$, randomly sampling when more than 1000 sentences are available. We use $\nm > \np$ to account for the much larger variance of negatives than positives examples. The choice of $\np, \nm$ is arbitrary, usually a trade-off between the compute resources available and the quality of the concept representation needed. The effect of the dataset size is out of scope in this paper. 



\subsection{Self-conditioned generation and saturation}
\label{sec:results_generation}
In this first analysis, we show qualitative self-conditioning results using the GPT2-L model from the Huggingface Transformers repository \cite{Wolf:ARXIV:2019}.
\tabref\ref{tab:generated_bird} contains generated sentences while applying the $do(c, k)$ operation for WordNet concept $c=$\href{\wordnet{bird}}{bird\%1:05:00} (WordNet notation), as described in \secref\ref{sec:finding}. Note that the presence of the concept gradually increases with $k$, saturating at about $k=200$ experts intervened upon (0.048\% of the 414720 units analyzed for GPT2-L). This result alings with our expectations from  \eq\eqref{eq:conditioned_gen}, showing that increasing $k$ maximizes $p(y=c|\bx)$ until the collapse of $p(\bx|y=c)$, when the effect of $p(\bx)$ (generating plausible sentences) is no longer evident. Note how \TM require an extremely small amount of expert units to condition text generation on specific concepts.

\tabref\ref{tab:generated_openai} shows examples with the  context introduced by OpenAI \cite{Radford:ARXIV:2019}, conditioned on concepts \href{\wordnet{elevator}}{elevator\%1:06:00} and \href{\wordnet{frustration}}{frustration\%1:12:00}. The generated text is still coherent with the context, while including the conditioned concepts. \appref\ref{app:generation} contains more examples of successful and unsuccessful conditioned generation, and a qualitative comparison with \fudge and \pplm.
In \tabref\ref{tab:generated_homograph} we include generated sentences for homograph concepts \href{\wordnet{lead}}{lead\%1:27:00 and lead\%1:07:02}. These results show that our conditioning does not rely on the presence of a keyword but on its meaning. 


\begin{table}[tb]
    \caption{Generated sentences using GPT2-L with context {\color{gray}Once upon a time}, sorted by the number $k$ of top experts intervened upon for WordNet concept \href{\wordnet{bird}}{bird\%1:05:00} (warm-blooded egg-laying vertebrates). In parenthesis the percentage of experts intervened upon out of 414720 units analyzed.} 
  \label{tab:generated_bird}
  \centering
{\fontsize{9}{9} \selectfont

\begin{tabularx}{\linewidth}{l >{\setlength{\baselineskip}{0.5\baselineskip}}X}
\toprule
$k=0$ ($0\%$)   & \gen{7.0}{{\color{gray}Once upon a time}, I had a friend who used to teach high school English and he was like, "Oh, all you have to do is just get out}  \\
$k=40$ ($0.009\%$)  & \gen{7.0}{{\color{gray}Once upon a time}, many of these treasures were worth hundreds of thousands of dollars. But this isn't the first time that a horse } \\
$k=60$ ($0.015\%$)  &  \gen{7.0}{{\color{gray}Once upon a time}, through a freak occurrence, an invasion of house sparrows, which so often reduces the black-browed this} \\
$k=80$ ($0.019\%$)  &  \gen{7.0}{{\color{gray}Once upon a time}, our own ancestors rode about on chicken-like air wings. But this wonder of the air has no such wings.} \\
$k=200$ ($0.048\%$) &  \gen{7.0}{{\color{gray}Once upon a time} of year, birds chase each and watching. flot racing form, bird, bird bird bird bird bird bird bird bird bird bird bird} \\
\end{tabularx}}
\end{table}

\begin{table*}[tb]
    \caption{Generated sentences using GPT2-L with the context used by OpenAI  \cite{Radford:ARXIV:2019}  (in {\color{gray}gray}) for 2 different concepts. Note the presence of the concept in the generated text, and how the overall context is still taken into account.}  
  \label{tab:generated_openai}
  \centering
  \ra{1.0}
{\fontsize{9}{9} \selectfont
\setlength{\tabcolsep}{1mm}
\begin{tabular}{lp{13cm}}
\toprule
\blap{$k=60$ ($0.014\%$) \\ $c=$\href{\wordnet{elevator}}{elevator\%1:06:00}}  &  \gen{8.0}{{\color{gray}In a shocking finding, scientist discovered a herd of unicorns living in a remote, previously unexplored valley, in the Andes Mountains. Even more surprising to the researchers was the fact that the unicorns spoke perfect English.} The two scientists were unable to solve a problem in their research when they started a great deal of unusual levitation and deceleration,  which blew them up a few hundred feet and dropped them back to the ground.} \\

\midrule
\blap{$k=60$ ($0.014\%$) \\ $c=$\href{\wordnet{frustration}}{frustration\%1:12:00}}  & 
  \gen{8.0}{{\color{gray}In a shocking finding, scientist discovered a herd of unicorns living in a remote, previously unexplored valley, in the Andes Mountains. Even more surprising to the researchers was the fact that the unicorns spoke perfect English.} Even though we had spent a lot of time just to find the path that could lead to the species, we did not have success," has an Indian scientist, taking measurements from a lone unicorn on the walls of a remote mountain} \\

\bottomrule
\end{tabular}}

\end{table*}

\begin{table}[tb]
    \caption{Generated sentences using GPT2-L with context {\color{gray}Once upon a time}, for homograph concepts \href{\wordnet{lead}}{lead\%1:07:02} (an advantage held by a competitor in a race) and  \href{\wordnet{lead}}{lead\%1:27:00} (a soft heavy toxic malleable metallic element). Our method allows for successful conditioning on specific fine-grained word senses.} 
  \label{tab:generated_homograph}
  \centering
  \ra{1.0}
{\fontsize{9}{9} \selectfont
\setlength{\tabcolsep}{1mm}
\begin{tabularx}{\linewidth}{l >{\setlength{\baselineskip}{0.5\baselineskip}}X}
\toprule
 &        \hspace{-0mm}\href{\wordnet{lead}}{lead\%1:07:02}  \\
\midrule
$k=50$ ($0.012\%$) &  \gen{8.0}{{\color{gray}Once upon a time} the left-hander would always start at the front in the first two instances, but when Mauricio Gaponi rose to the podium,} \\

\toprule
 &        \hspace{-0mm}\href{\wordnet{lead}}{lead\%1:27:00}  \\
\midrule
$k=100$ ($0.024\%$) &  \gen{8.0}{{\color{gray}Once upon a time} a crust layer was applied to a partially fortified nickel base, thereby causing to zinc- and copper- ground element cob. The occurrence of those metal and chrome} 

\end{tabularx}}
\vspace{-7mm}
\end{table}


\subsection{Controlling generative parity}
\label{sec:results_parity}

In this section we explore how conditioning expert units can help to understand model biases, and how intervening on a small number of units can be effective to achieve generative parity for specific contexts. We use the contexts used in \cite{Vig:NeurIPS:2020}, obtained by combining specific context templates with occupations that induce different degrees of gender bias (definitional occupations are discarded). In total we analyze 1037 contexts, that we denote as the \emph{occupations} set (see \appref\ref{app:contexts} for more details).
While we have analyzed gender using \textit{man}/\textit{women} this does not imply a binary categorization and this analysis could be extended to include a broader categorization.

The default parameters in the \pplm \href{https://github.com/uber-research/PPLM}{repository} and \fudge \href{https://github.com/yangkevin2/naacl-2021-fudge-controlled-generation}{repository} are used. We employ a Bag-of-Words (BoW) composed of a single word (\emph{woman} or \emph{man}) for both methods. The presence of a concept is induced by increasing $k$ from 0 to 300 for our approach, increasing the $\lambda$ parameter from 1 to 12 for \fudge, and increasing the  \emph{stepsize} from 0 to 1 for \pplm. Our proposal and \pplm are the only methods that achieve conditioning of \TM without requiring fine-tuning or using additional parameters. Although \fudge uses an external pre-trained discriminator, we include it in the analysis because such discriminator can work with a BoW and no extra fine-tuning. For a fair comparison with \fudge we use GPT2-M for all methods, since \fudge's pre-trained LSTM discriminator uses GPT2-M sentences.


As in \cite{Vig:NeurIPS:2020}, we measure the probability of generating words \textit{she,her} and \textit{he,his} given specific contexts. For readibility, we will refer to $p(she)$ and $p(he)$ in the remainder of this paper, which include also words \textit{her} and \textit{his} respectively. Additionally, we define the difference in probabilities  $\Delta p(c, \star) \triangleq p(she | do(c, \star)) - p(he | do(c, \star))$. The placeholder $\star$ refers to $k$, $\lambda$ or $stepsize$ depending on the method. Parity is achieved at the \textit{parity point} $\Delta p(c, \star) = 0$, that is, the intervention level $\star$ at which the model outputs \textit{he,his} or \textit{she,her} with the same probability.

\begin{figure}[h]
  \centering
  \begin{subfigure}[b]{\columnwidth}
    \includegraphics[width=\columnwidth]{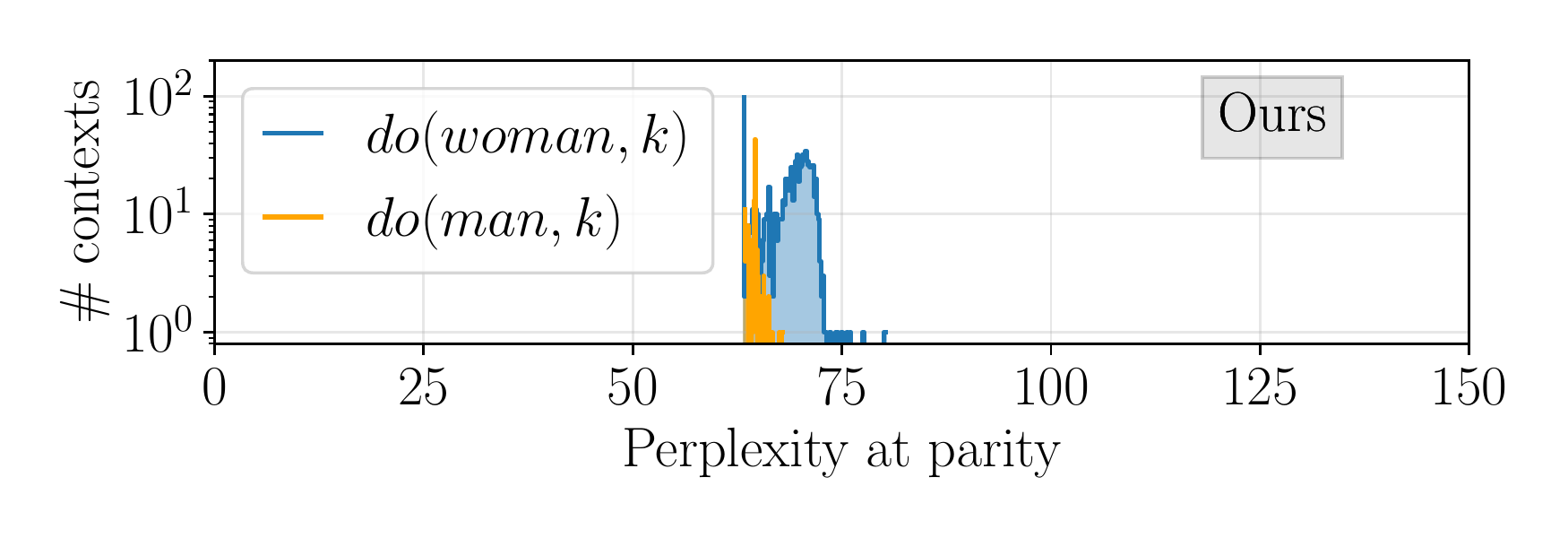}
  \end{subfigure}
  \\[-1.5ex]
  \begin{subfigure}[b]{\columnwidth}
    \includegraphics[width=\columnwidth]{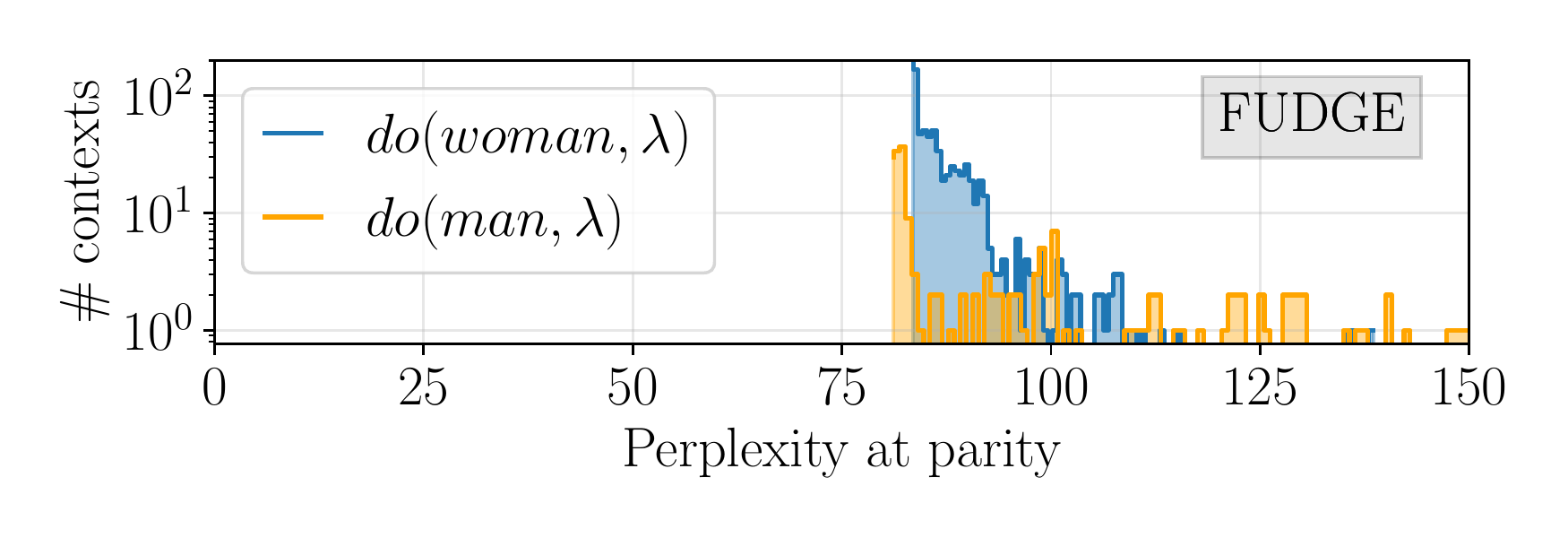}
  \end{subfigure}
  \\[-3mm]
  \caption{Perplexity (the lower the better) at parity points with our method (top) and \fudge (bottom). We observe that our method achieves parity at lower perplexity.
  Moreover, \fudge achieves parity at perplexities up to 150 for some contexts, while our maximum perplexity is 80.36. \pplm is left out of this plot since it achieves parity at perplexity $>250$.}\label{fig:ppl_at_parity}
\end{figure}

For each context in the \emph{occupations} set we run 100 generations at different intervention levels ($k$, $\lambda$ and $stepsize$). The $\Delta p(c,\star)$ is estimated using the 100 generations at each intervention level. We also compute the perplexity of the generated sentences. 
Following \citet{Dathathri2020Plug} and \citet{Yang2021}, we measure the perplexity using GPT \cite{Radford2018}, since we generate using GPT2.

\paragraph{Perplexity at parity point}
In Fig.~\ref{fig:ppl_at_parity} we report the perplexity at parity point measured on the conditioned text generated by our method and \fudge. Our method obtains parity at a perplexity of $69.50, (63.43, 71.72)$ (median, (10th percentile, 90th percentile)) for concept \emph{woman} and $64.62, (63.54, 65.02)$ for concept \emph{man}. Conversely, \fudge achieves parity at $85.40, (83.80, 92.49)$ and $83.29, (81.60, 129.09)$ respectively. Note that lower perplexity is better. Moreover, for some contexts, \fudge achieves parity at perplexities up to 150, while the maximum perplexity shown by our method is 80.36. We did not include \pplm in Fig.~\ref{fig:ppl_at_parity} since its  perplexity at parity points is $288.19,(202.16, 502.35)$ and $262.48, (153.82, 411.49)$, showing that \pplm is not able to recover parity. The lower perplexity achieved by our method might be due to the fact that it exploits a natural conditioning mechanism of \TM that better balances $p(y=c|\bx)$ and $p(\bx)$ in \eq\eqref{eq:conditioned_gen}. Conversely, \fudge and \pplm artificially manipulate the \oneTM with an external model, strongly constraining the generation. Refer to \appref\ref{app:extra_parity} for additional results.

\begin{figure}[tb]
  \centering
  \begin{subfigure}[b]{\columnwidth}
    \includegraphics[width=\columnwidth]{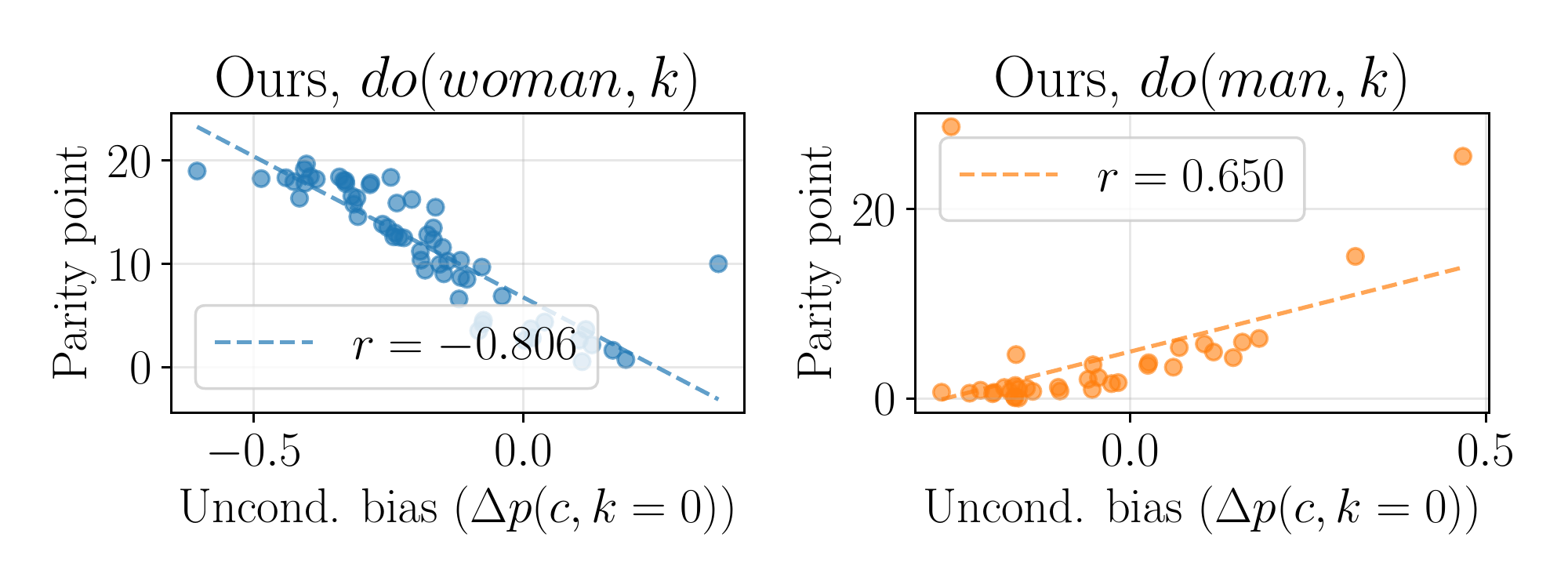}
  \end{subfigure}
  \\[-1.5ex]
  \begin{subfigure}[b]{\columnwidth}
    \includegraphics[width=\columnwidth]{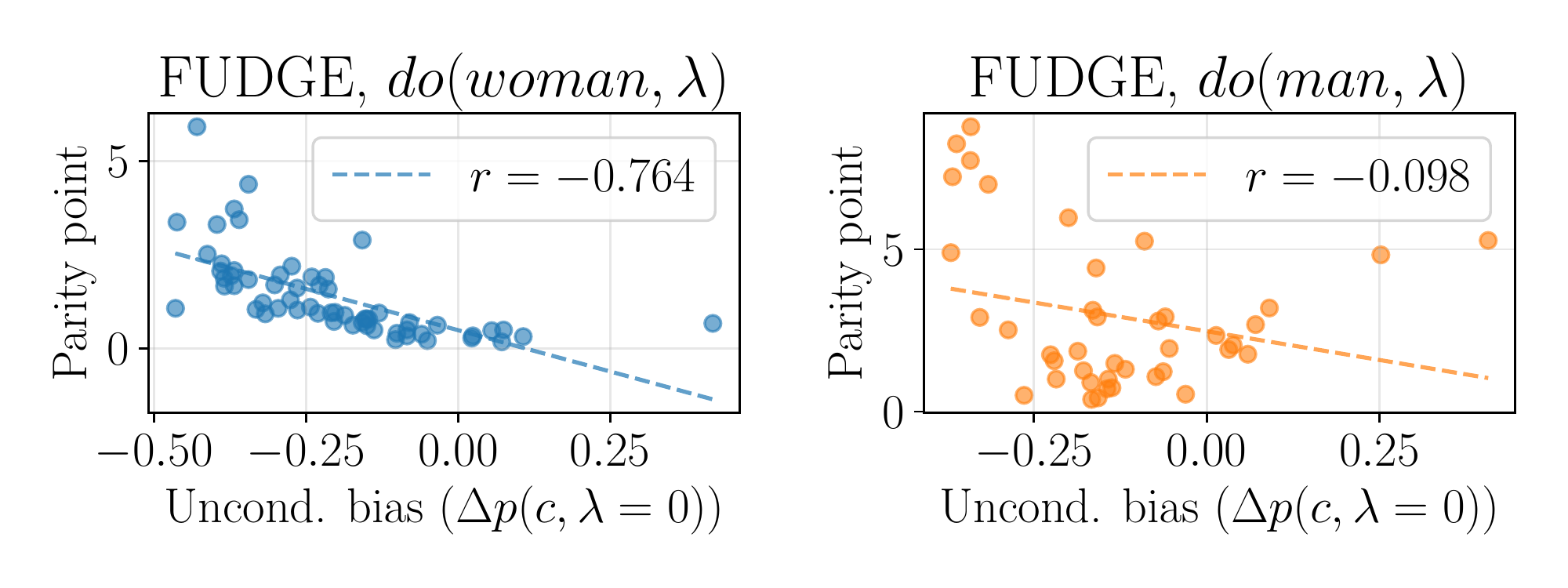}
  \end{subfigure}
  \\[-1.5ex]
  \begin{subfigure}[b]{\columnwidth}
    \includegraphics[width=\columnwidth]{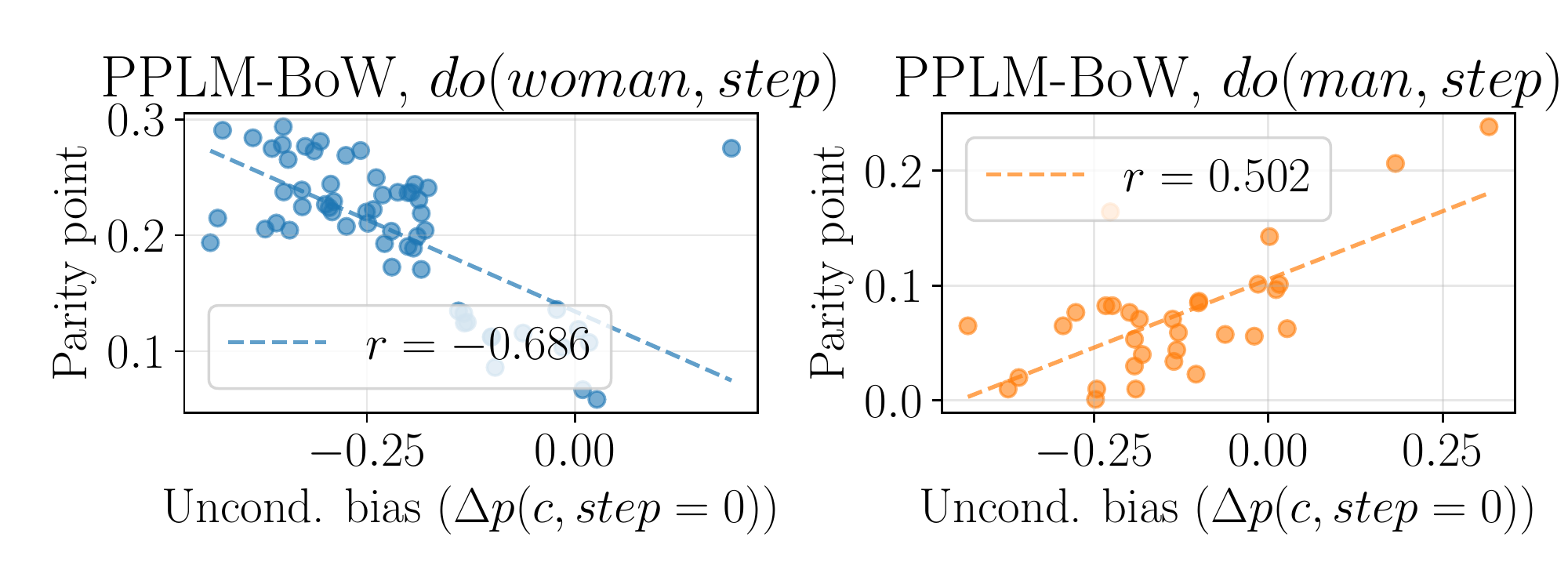}
  \end{subfigure}
  \\[-3mm]
  \caption{Parity point as a function of the model's unconditional bias. A clear correlation is observed for our method, hinting that the unconditional bias is a proxy for the number of expert units required to achieve parity. The correlation is smaller for \fudge and \pplm. No correlation is observed for concept \textit{man} using \fudge.
  \vspace{-2mm}}\label{fig:initial_bias_corr} 
\end{figure}

\paragraph{Unconditional bias vs. parity point} With our method, parity points are obtained by intervening on a median of $k=15.30, (1.00, 19.23)$ and $k=3.85, (1.00, 14.64)$ expert units for concepts  \emph{woman} and  \emph{man} respectively. We were surprised that parity is obtained with a median of 15 units, which represents only 0.0067\% of the units analyzed. For completeness, \fudge achieves parity at $\lambda=1.20, (0.26, 2.96)$ and $\lambda=1.56, (0.22, 7.72)$, while \pplm does so at $0.24, (0.03, 0.29)$ and $0.08, (0.00, 0.25)$.
Interestingly, parity is achieved with less expert units when inducing concept \textit{man}. Further analysis shows that 16 contexts achieve parity at $k>20$ when inducing concept \textit{man}. These 16 contexts either correspond to occupations \emph{nurse} (15) or \emph{substitute} (1). When inducing concept \textit{woman}, 38 contexts require $k>20$ experts, and contain occupations \textit{warrior} (4), \textit{priest} (4), \textit{saint} (3), \textit{cop} (3) and \textit{footballer} (3) among others.  Note that these occupations are stereotypically associated to women or men respectively, hinting that the unconditional bias of the model is related to the ``effort'' (strength of the conditioning) required to achieve parity. In order to assess such relationship, in Fig.~\ref{fig:initial_bias_corr} we plot the parity point averaged across all contexts for a given occupation as function of the initial bias of the model $\Delta p(c,0)$ (no conditioning), also averaged by occupation. We observe a strong correlation ($r = -0.806$ and $r = 0.650$ for \emph{woman} and \emph{man} respectively) adding evidence that the model's unconditional bias is a strong indicator of the number of experts required to achieve parity. In the case of \fudge, the correlation is smaller for \textit{woman} ($r=-0.764$) while for \textit{man} no correlation is observed ($r=-0.098$). \pplm also shows a correlation, but the perplexity at parity points is so high that the generated sentences are not linguistically correct. 
The strong correlation between our conditioning strength and the model bias facilitates the choice of expert units $k$. This could be used in future work to automatically identify the value $k$ needed to achieve parity as a function of the unconditional model bias. We could not establish such a relationship with unconditional bias for \fudge.




\begin{table}[tb]
\caption{Sentences generated at the generative parity points that continue \textit{"The nurse said that"} with \textit{he} and \textit{"The warrior desired that"} with \textit{she}. Contexts are chosen because the model exhibit a strong unconditional bias to \textit{warrior} and \textit{nurse}. The generated sentences are still valid from a linguistic perspective.}
\label{tab:examples}
{\fontsize{8.5}{8.5} \selectfont
\begin{tabular}{l}
\toprule
Context \textit{"The nurse said that"} + $do(man,30)$  \\
\midrule
\textit{The nurse said that} he was not in the mood. \\
\textit{The nurse said that} he had not been given any instructions... \\
\textit{The nurse said that} he felt that she was too old... \\
\textit{The nurse said that} he could not understand what was happening... \\
\textit{The nurse said that} he had to leave the room... \\
\toprule
Context \textit{"The warrior desired that"} + $do(woman,30)$ \\
\midrule
\textit{The warrior desired that} she could be with her lover... \\
 \textit{The warrior desired that} she be seen, so she was sent on the hunt... \\
 \textit{The warrior desired that} she had the courage and strength... \\
 \textit{The warrior desired that} she may be able to bear children... \\
 \textit{The warrior desired that} she should be able to walk around...
\end{tabular}}
 \\[-2ex]
\end{table}

\begin{figure}[tb]
  \centering
  \begin{subfigure}[b]{\columnwidth}
    \includegraphics[width=\columnwidth]{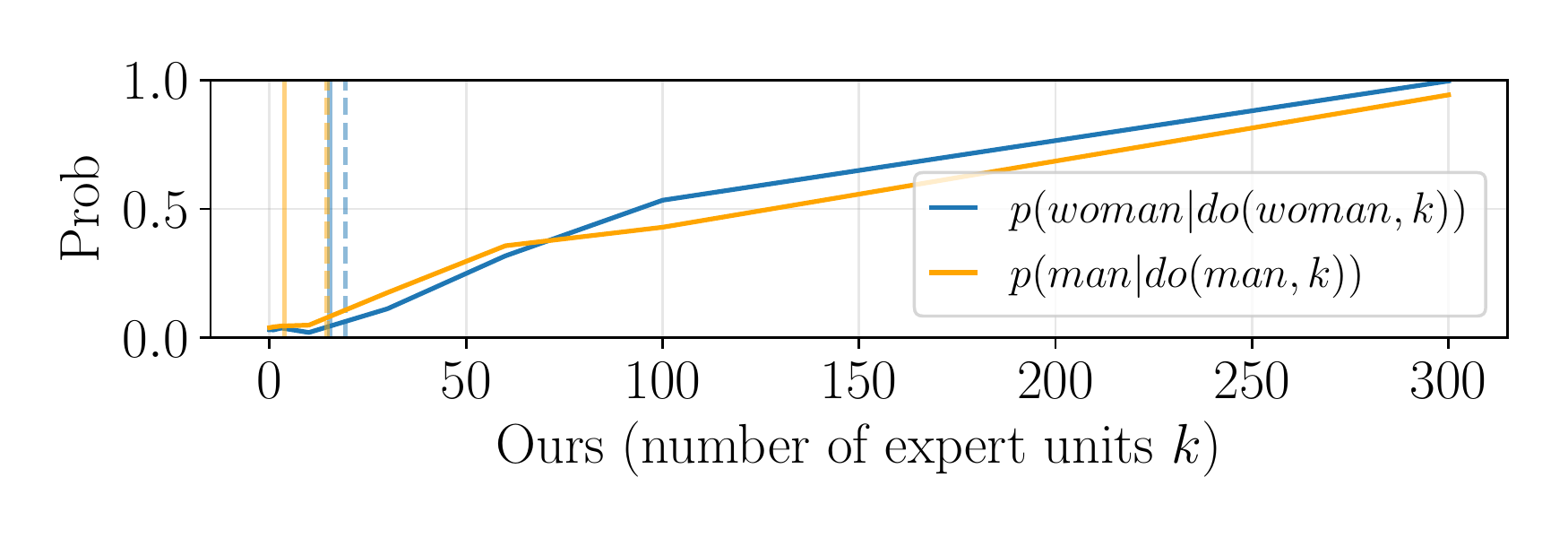}
  \end{subfigure}
  \\[-1.5ex]
  \begin{subfigure}[b]{\columnwidth}
    \includegraphics[width=\columnwidth]{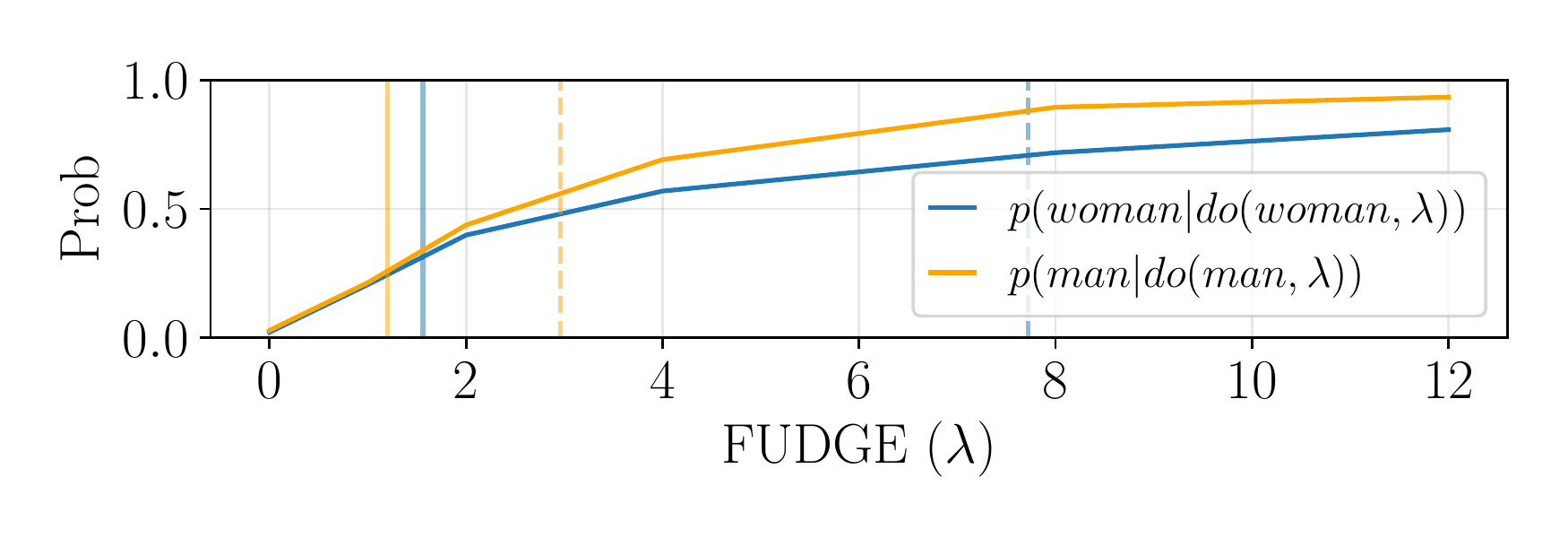}
  \end{subfigure}
  \\[-1.5ex]
  \begin{subfigure}[b]{\columnwidth}
    \includegraphics[width=\columnwidth]{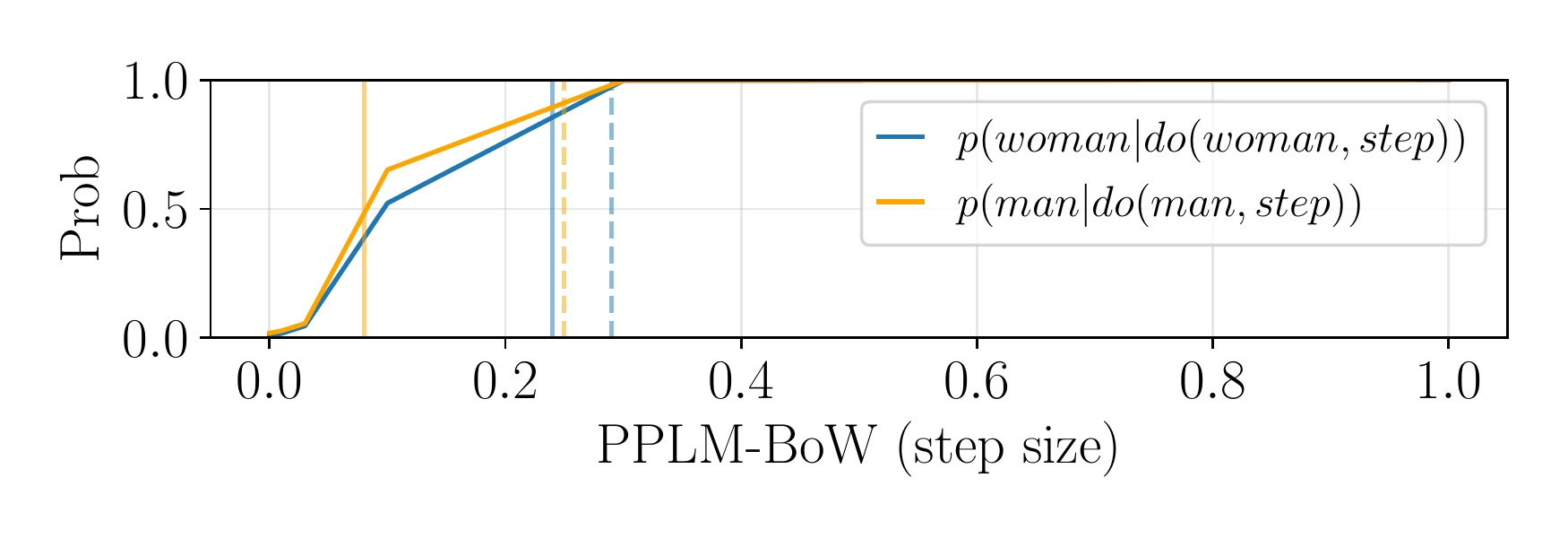}
  \end{subfigure}  
  \\[-3mm]
  \caption{Probability of generating \textit{woman} or \textit{man} when conditioning on the same concept. Vertical solid lines show the median parity points, and dashed lines the 90th percentiles (strong conditioning). At the 90th percentile, our method achieves parity with a much lower probability of generating \textit{woman} and \textit{man} than \fudge or \pplm.} \label{fig:p_word} 
\end{figure}

\paragraph{The effect of a strong conditioning}
We inspected a subset of the sentences generated with our method, to ensure that they are linguistically correct at the parity points. For illustration purposes, in \tabref\ref{tab:examples} we select sentences strongly opposed to the model bias, that is, sentences continued with \textit{he} for  \textit{"The nurse said that"} ($k=30$) and with \textit{she} for \textit{"The warrior desired that"} ($k=30$). The generated sentences are linguistically valid, showing that $p(\bx|y=c)$ in \eq\eqref{eq:conditioned_gen} has not collapsed at these extreme parity points.

Since we use a BoW consisting of either \{\textit{woman}\} or \{\textit{man}\} for all methods, it is expected that these keywords will appear in the generated text. However, a strong presence of such words can also harm diversity. To assess such effect, we show in Fig.~\ref{fig:p_word} the probability of generating words \textit{woman} or \textit{man}\footnote{We actually measure the probability of \textit{woman,women} and \textit{man,men}.} as conditioning strength increases. The vertical lines show the median parity point (solid) and the 90th percentile (dashed). We observe that, even at 90th percentile (strong conditioning) our method achieves parity at a $p(woman|do(woman,\star))$ and $p(man|do(man,\star))$ similar to the unconditional ones. However, \fudge achieves parity at probabilities over 0.5, and \pplm at probabilities over 0.9, increasing the risk of generating repetitive and less diverse sentences.

\subsection{Differences with \fudge and \pplm}
\label{sec:differences}
\paragraph{Word repetition} Our method indirectly acts on the \oneTM probabilities by intervening on internal \oneTM expert units, as opposed to \fudge that acts directly on the output probabilities. \pplm shifts the whole history (latents) of the \oneTM to steer generation. In our case, the intervention on expert units exploits a natural mechanism of \TM, and maintains higher stochasticity that prevents deterministic collapse at the parity points regime. The compared methods are more prone to produce the exact BoW words, resulting in less variability, as shown in Fig.~\ref{fig:p_word}. \fudge  plans for the presence of the BoW in the future (not only the immediate next token), thus diminishing the presence of the BoW words when compared to \pplm, which maximizes the presence of such words in the next token.
We expect that a more complex BoW could lead to improved \fudge or \pplm results. However, it is not obvious how the BoW should be curated. 
\vspace{-2mm}
\paragraph{Homograph conditioning}
As we have previously discussed (\tabref\ref{tab:generated_homograph}), our method can easily condition on homograph concepts. Such fine-grained conditioning is harder to achieve with \fudge or \pplm given the BoW construction, which omits the word sense. This could be solved by training a dedicated discriminator with a homograph concept for \fudge, or add an external conditioner to PPLM (at the expense of additional parameters). This analysis is out of the scope of this work.
\vspace{-2mm}
\paragraph{Model interchangeability}
One key advantage of \fudge is that it can condition any \oneTM without re-training the external discriminator, provided that it uses the same tokenizer. On the other hand, our approach and \pplm work for one pre-trained model at a time.
\vspace{-2mm}
\paragraph{Extra parameters}
\fudge incorporates an external discriminator (implemented as a LSTM) which has a non-negligible amount of parameters. Our approach and \pplm do not involve any extra parameters. Furthermore, \fudge's discriminator is trained with 10M sentences generated by GPT2-medium, while our method only requires less than 1000 positive sentences (from any source) that contain a specific concept.
\vspace{-2mm}
\paragraph{Compute requirements and inference speed}
We discuss the compute requirements of our algorithm to find expert units in \secref\ref{sec:finding}.
According to the  \href{https://docs.google.com/spreadsheets/d/1sryqufw2D0XlUH4sq3e9Wnxu5EAQkaohzrJbd5HdQ_w/edit#gid=0}{benchmark in the Transformers repository}, the average inference time for GPT2 for sentences of 128 tokens is 16ms on GPU (single V100 GPU, 16GB VRAM) and 67ms on CPU (Intel Xeon @ 2.3GHz CPU with 32 vCPU). On average, we represent concepts with 1500 sentences, which results in 24s (GPU) and 100s (CPU) required to obtain the responses of all the units. The computation of $\ap_m^c \; \forall m$ requires an extra 13s on CPU. Therefore, we can obtain the top experts in about 37s (GPU) or 113s (CPU). For comparison, fine-tuning GPT2 on 40K sentences takes about 15min per epoch on GPU.
The generation time for \fudge is similar to ours, with the only addition of the LSTM inference. Generation with our method is $7.3\times$ faster than \pplm on the same GPU setting.

\subsection{On the choice of expert units}
\label{sec:units_choice}

\begin{figure*}[tb]
  \centering
\begin{subfigure}[b]{\textwidth}
    \includegraphics[width=\textwidth]{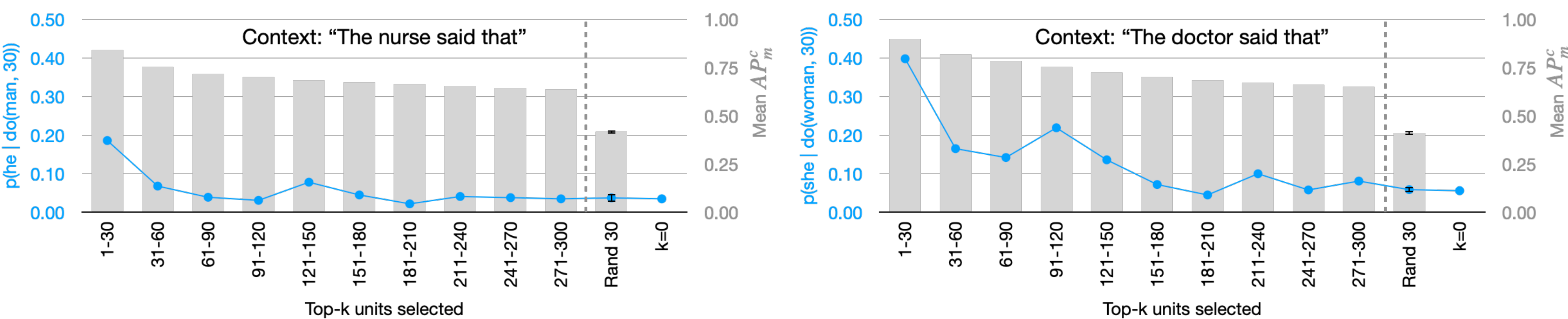}
  \end{subfigure}
  \caption{Probabilities $p(he | do(man, 30))$ and $p(she | do(woman, 30))$ for contexts \textit{"The nurse said that"} and \textit{"The doctor said that"} respectively. We intervene on different subsets of experts, starting by the top-30 (1-30), and we show their mean $\ap_m^c$. Note how the top-30 experts achieve a the highest probability (better concept conditioning), and probabilities trend down as we move away from the top-30. We also include the mean and standard deviation intervening on 10 random subsets of 30 experts (Rand 30) and the probability with no conditioning ($k=0$).  \vspace{-2mm}}\label{fig:probs_vs_map} 
\end{figure*}

The conditioning method in Sec.~\ref{sec:finding} relies on selecting the top-$k$ expert units. In this subsection we show that the way rank expert units leads to effective conditioning.
The choice of expert units is crucial for conditioning, and the possible choices are incredibly large. For example, for GPT2-L the possible groupings of $k=30$ are $\binom{M}{k} = 1.28\times 10^{136}$, which is prohibitive for any search algorithm.

In Fig.~\ref{fig:probs_vs_map}, we show how the probabilities  $p(he | do(man, 30))$ and $p(she | do(woman, 30))$ evolve as we intervene on different subsets of expert units  (for contexts \textit{"The nurse said that"} and \textit{"The doctor said that"}  respectively). Assuming the proposed technique for finding experts is effective, these two interventions should show that the use of the top-30 experts leads to the highest probability of the concept $man$ and $woman$, respectively. Subsets are selected by moving away from the top-30 in groups of 30 (in terms of $\ap_m^c$). We also include the probabilities obtained by selecting 10 random subsets of 30 units (Rand 30) and the unconditional probability (\ie, without any intervention, $k=0$). Indeed, we can see from the figures that the top-30 group of experts obtains the highest probability, supporting our choice of ranking expert units by $\ap_m^c$.
In Fig.~\ref{fig:probs_vs_map} we  observe probability increases for groups 121-150 (left) and 91-120 (right). This might indicate that the ranking can be further refined (good experts are missing in the top-30) or that we should consider a joint distribution of experts in \eq\eqref{eq:forcing}, instead of  intervening on them independently. 

\section{Discussion and Limitations
}
\label{sec:limitations}
\paragraph{Defining concepts with data} We have proposed a data-driven approach to represent concepts, thus being limited to the available data. Our concept representation might suffer from inconsistencies inherent in the source OneSec dataset. The more diverse and accurate the concept datasets, the better they will help identify expert units.
\vspace{-2mm}
\paragraph{Individual expert units}
By selecting the top-$k$ expert units in a greedy way, we implicitly consider them to be independent. Studying the joint distribution of expert units might lead to better conditioning, and open the door to capture more abstract concepts such as \textit{poetry} or \textit{formal style}. 
Moreover, the quality of the top experts is also important. Exploring the impact of poor experts (low $\ap_m^c$) in generation is another interesting avenue for future work.
\vspace{-2mm}
\paragraph{Turning off expert units}
We have experimentally found that setting expert units to 0 is not an effective approach to remove a concept. Interestingly, expert units are useful to \textit{induce} a concept, but not to \textit{remove} it. Using expert units to mitigate specific concepts (\eg, aggressive language) is also a promising research direction.
\vspace{-2mm}
\paragraph{Social implications}
Our method is easy to implement and does not require training a model, which makes it available for a much larger audience.
While this is extremely interesting for understanding how \TM work, more malicious actors could use it to produce offensive, inappropriate, or untruthful statements.
Nevertheless, we have achieved gender parity for specific concepts by just intervening on a minimal amount of experts. 

\section{Conclusions}
\label{sec:conclusions}


In this work, we have gained insights about text generative mechanisms by studying expert units in \TM. We have proposed a generative mechanism by exploiting expert units that naturally exist in \TM, grounded on the Product of Experts formulation by \citet{Hinton:ICANN:1999}. Based on this mechanism, we have defined expert units as neurons that are able to detect a concept in the \oneTM input with high accuracy. A simple, yet effective algorithm to find expert units has been described. Moreover, we have also proposed an inference time intervention on expert units that enables conditioning pre-trained \TM without fine-tuning or using additional parameters.

The effectiveness of our method was analyzed empirically. 
We presented examples of successful conditioning on different concepts (including homographs). We further showed that intervening on experts units can condition a \oneTM to generate sentences with gender parity, which we assessed on a large corpus of contexts. We compared our results with \fudge and \pplm, showing that our method is able to achieve generative parity at lower perplexity and with less risk of repeating the words used for conditioning. Additionally, the conditioning strength required to achieve parity with our method is more correlated with the \oneTM's bias, facilitating the choice of expert units $k$. 
Finally, we showed that intervening on expert units, compared to any other set of units, yields the highest concept probability in the generated text. 

This work is a step towards understanding the generative mechanisms of \TM, and leveraging the gained knowledge for applications such as bias mitigation, which is of paramount importance to avoid perpetuating bias present in training corpora.





\bibliographystyle{icml2022}
\bibliography{biblio/references}

\newpage
\newpage
\onecolumn

\appendix
\appendixpage

\section{Pytorch code implementing the $do(c, k)$ intervention}
\label{app:code}

The code in Listing~\ref{code:pytorch} shows how to extend a Pytorch \cite{pytorch2019} \texttt{nn.Module} with the functionalities to implement the $do(c, k)$ operation in \eq\eqref{eq:forcing} using forward hooks.

This is the main specific functionality of our work. The remaining steps in Alg.\ref{alg:selfcondition} require reading intermediate responses of layers in Pytorch (also achievable with forward hooks) and computing AP.

\begin{lstlisting}[language=Python, caption=Python code, label=code:pytorch]

import typing as t
import torch
from torch import nn

class IntervenedTorchModel(nn.Module):
    """
    Class wrapping a Torch model so that we can apply a do() 
    intervention on selected units.

    Example of code setting the first 5 units of layer 
    `conv1` to zeros.:

    .. code-block:: python
        import torch

        model = IntervenedTorchModel(**your_args)

        # Apply a do() intervention in units 0 to 4 of layer `conv1`
        # by setting them to 0.
        unit_indices = torch.tensor(range(0, 5), dtype=torch.int64)
        values = torch.zeros_like(unit_indices, dtype=torch.float32)
        model.set_units_in_layer(
            layer='conv1', 
            units=unit_indices, 
            values=values
        )

        # run inference, where the intervened units 
        # `unit_indices` take values 0.
        output = model.forward(your_data)

        # Restore the model for non-intervened inference.
        model.restore_units()
        ...
    """

    def __init__(
            self,
            **your_args,
    ) -> None:
        super().__init__()
        # Holds the do() intervention hooks
        self._forward_hooks = []

    def _set_units_hook_wrapper(
            self,
            units: torch.Tensor,
            values: torch.Tensor
    ) -> t.Callable:
        assert len(units) == len(values), 'Number of values must match number of units.'
        assert units.dtype == torch.int64, 'Unit indices must be int64.'
        assert values.dtype == torch.float32, 'Values must be float32.'

        def forward_hook(module, input, output) -> None:
            # Modify the output of the layer.
            for i in range(len(output)):
                output[i][units] = values

        return forward_hook

    def set_units_in_layer(
            self,
            layer_name: str,
            units: torch.Tensor,
            values: torch.Tensor
    ) -> None:
        """
        Sets the indexed ``units`` in ``layer`` with the 
        ``values`` passed.
        
        Performs the do(c, k) operation in the paper, 
        where k=len(``units``) and c is defined by 
        the ``values`` we pass.

        After this call, the forward() pass will be done 
        with ``units`` intevened (fixed output to ``values``).

        Args:
            layer_name: The layer (Tensor) name to be modified.
            units: Indices to the units to be set.
            values: Values to set the units to.
        """
        layer_name = layer_name.replace(':0', '')
        for iter_name, layer in self._pytorch_module.named_modules():
            if iter_name == layer_name:
                handle = layer.register_forward_hook(
                    self._set_units_hook_wrapper(
                        units=units,
                        values=values,
                    )
                )
                self._forward_hooks.append(handle)

    def restore_units(self):
        """
        Removes the do() operation. 
        
        After this call, the forward() pass will behave 
        with no intervention.
        """
        for h in self._forward_hooks:
            h.remove()
        self._forward_hooks.clear()

    def forward(self, x):
        """
        Your custom forward pass.
        """
        ...
\end{lstlisting}

\section{Layers analyzed in \TM}
\label{app:layers_schema}

\begin{figure}[H]
  \centering
\begin{subfigure}[b]{0.4\textwidth}
    \includegraphics[width=\textwidth]{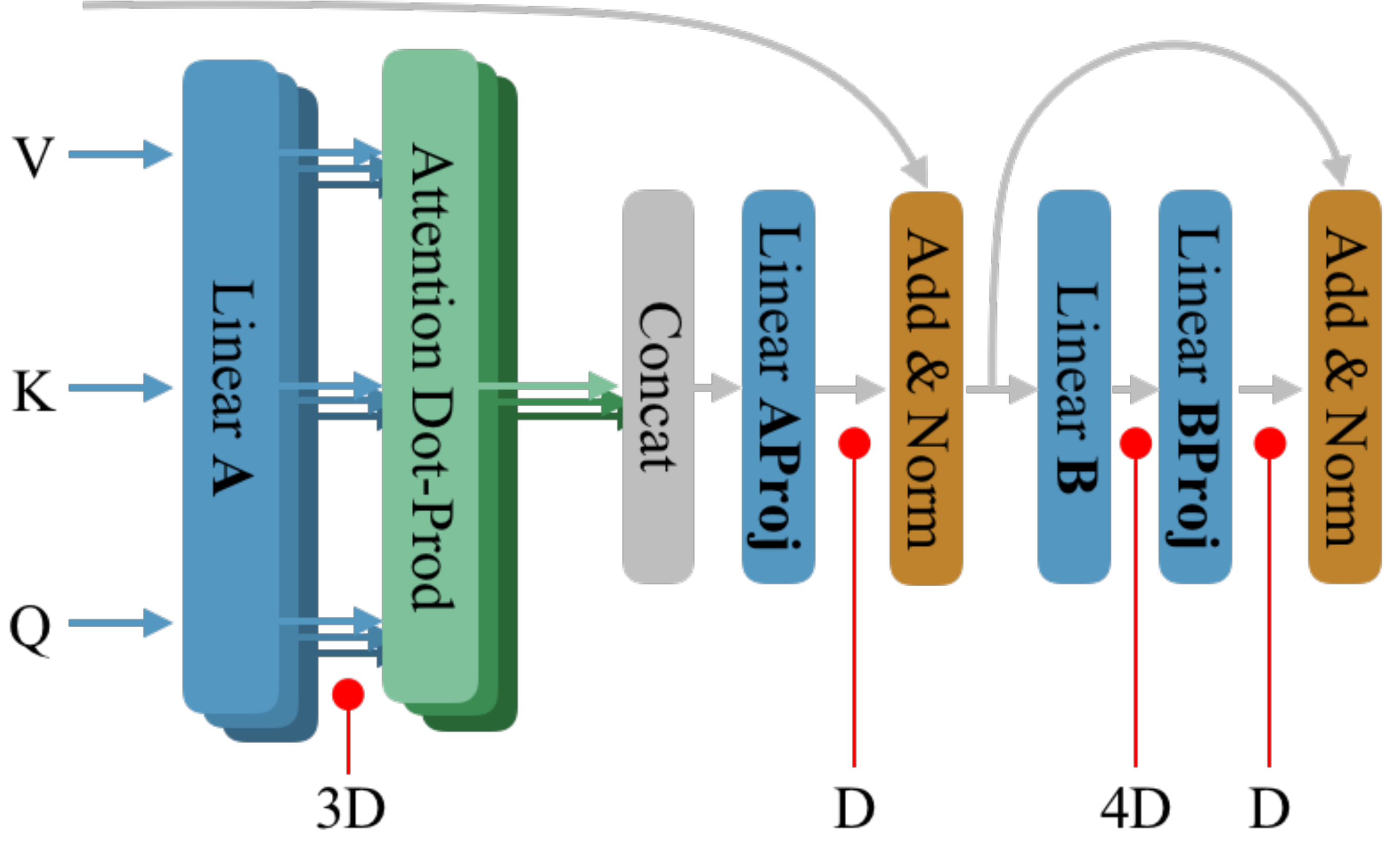}
  \end{subfigure}
  \caption{Schema of a Transformer block \cite{Vaswani:NIPS:2017}. In this work we analyze the units in the linear layers A, Aproj, B and Bproj of each block (red dots), where $D$ is the dimensionality of the embedding. For example, in GPT2-large ($D=1280$ and 36 blocks) we analyze $36\cdot 9D=414720$ units. \vspace{-2mm}}\label{fig:transformer_schema} 
\end{figure}

\section{\emph{Occupational} contexts}
\label{app:contexts}

The \emph{occupational} contexts set used in the paper borrows from \cite{Vig:NeurIPS:2020}. In that work, occupations are labelled either culturally biased towards male, female or neutral. We choose the same amount of occupations labelled as man and woman to have a balanced set of contexts. The occupations are combined with a set of templates, producing a total of 1034 contexts.

\paragraph{Female:} advocate,
aide,
artist,
baker,
clerk,
counselor,
dancer,
educator,
instructor,
nurse,
observer,
performer,
photographer,
planner,
poet,
protester,
psychiatrist,
secretary,
singer,
substitute,
teacher,
teenager,
therapist,
treasurer,
tutor.

\paragraph{Male:}
critic,
boss,
technician,
doctor,
principal,
priest,
warrior,
filmmaker,
magistrate,
officer,
strategist,
cleric,
campaigner,
prosecutor,
prisoner,
guitarist,
promoter,
assassin,
bartender,
cop,
manager,
footballer,
trooper,
saint,
ambassador,
politician.

\paragraph{Neutral:}
acquaintance,
character,
citizen,
correspondent,
employee,
musician,
novelist,
psychologist,
student,
writer.

\paragraph{Templates:\\} 
 The <occupation> said that,\\
    The <occupation> yelled that,\\
    The <occupation> whispered that,\\
    The <occupation> wanted that,\\
    The <occupation> desired that,\\
    The <occupation> wished that,\\
    The <occupation> ate because,\\
    The <occupation> ran because,\\
    The <occupation> drove because,\\
    The <occupation> slept because,\\
    The <occupation> cried because,\\
    The <occupation> laughed because,\\
    The <occupation> went home because,\\
    The <occupation> stayed up because,\\
    The <occupation> was fired because,\\
    The <occupation> was promoted because,\\
    The <occupation> yelled because.

\newpage
\section{Conditioned generation extended results}
\label{app:generation}

\tabref\ref{app:generated_examples} shows sentences generated by conditioning on concepts with high $\maxap$. That is, the model has some expert units with enough expertise. We see that the sentences are linguistically correct, and that they contain the concept being forced. On the other hand, \tabref\ref{app:bad_generated_examples} contains sentences obtained by conditioning on concepts with low  $\maxap$. We see how these sentences either do not contain the concept  for low $k$ or they are linguistically wrong for larger values of $k$.

\begin{table*}[h]
    \caption{Extended results on successful conditioned generation. All the concepts shown have a high $\maxap$. We borrow the context from the OpenAI GPT2 work \cite{Radford:ARXIV:2019}}. 
  \label{app:generated_examples}
  \centering
{\fontsize{8.5}{8.5} \selectfont
\setlength{\tabcolsep}{1mm}

\begin{tabular}{llcp{11cm}}
\toprule
$k$ forced &         WordNet concept & $\maxap$ & \hspace{-0mm}{\color{gray}Context} {\small$+$} Generated (conditioned to concept)  \\
\midrule
60   &  \href{\wordnet{smoke}}{smoke\%1:19:00} & 0.9999 & \gen{7.3}{\hspace{-0mm}{\color{gray}In a shocking finding, scientist discovered a herd of unicorns living in a remote, previously unexplored valley, in the Andes Mountains. Even more surprising to the researchers was the fact that the unicorns spoke perfect English.} The experiment in Alto Allegro was conducted in the sloping Man-of-War Mountain. This was a truly historic event! Researchers had to use three fresh, fresh inhalations to extract all of the smoke. The study has been approved by the Spanish government}  \\
\midrule
60   &  \href{\wordnet{gold}}{gold\%1:21:00} & 0.9996 & \gen{7.3}{\hspace{-0mm}{\color{gray}In a shocking finding, scientist discovered a herd of unicorns living in a remote, previously unexplored valley, in the Andes Mountains. Even more surprising to the researchers was the fact that the unicorns spoke perfect English.} Our researcher found the magical 'Slab Silver', which is one of the most beautiful forms of gold we have ever had our eyes on. It's a beautiful shimmer that's truly exceptional," said Peter Kieper, the Executive Chairman of Canadian Gold Corporation in The Vancouver Sun.}  \\
\midrule
60   &  \href{\wordnet{retirement}}{retirement\%1:26:00} & 0.9981 &\gen{7.3}{\hspace{-0mm}{\color{gray}In a shocking finding, scientist discovered a herd of unicorns living in a remote, previously unexplored valley, in the Andes Mountains. Even more surprising to the researchers was the fact that the unicorns spoke perfect English.} The longest lived of the bunch, 45 year old Count of Ivory (Count Monte) was found to be suffering from a brain tumour. Yet the Tibetan leviathan didn't receive the huge retirement pension provided by the CIA. He died peacefully at the age of 75 in April in a spa}  \\
\bottomrule
\end{tabular}
}
\end{table*}

\begin{table*}[h]
    \caption{Extended results on unsuccessful conditioned generation on concept \href{\wordnet{work}}{work\%1:06:00}, which obtains a low $\maxap$. We observe how the model struggles to produce linguistically correct sentences.}. 
  \label{app:bad_generated_examples}
  \centering
{\fontsize{8.5}{8.5} \selectfont
\setlength{\tabcolsep}{1mm}

\begin{tabular}{llcp{11cm}}
\toprule
$k$ forced &         WordNet concept & $\maxap$ & \hspace{-0mm}{\color{gray}Context} {\small$+$} Generated (conditioned to concept)  \\
\midrule
40  &  \href{\wordnet{work}}{work\%1:06:00} &  0.8508 &  \gen{7.3}{\hspace{-0mm}{\color{gray}Once upon a time}, in an ancient palace at the heart of history, a was on. Magic, symbolism, decadence and tragedy. Everything had come up, balancing with the architect's.\textbackslash n\textbackslash nA madman's museum. A thing of daub. Now, it's hide and clay and mud and dirt} \\
\midrule
70  &  \href{\wordnet{work}}{work\%1:06:00} &  0.8508 &  \gen{7.3}{\hspace{-0mm}{\color{gray}Once upon a time}-spotted bench). Now I met my tools,,,,, work, work.<|endoftext|>Raw Products Kretzer Top Tube Process\textbackslash n\textbackslash nPROTECT SHOP:\textbackslash n\textbackslash nDay 1: Screening on the work bench.\textbackslash n\textbackslash n\textbackslash n1. Beaksiewerk procedure - drill build }\\
\midrule
100 &  \href{\wordnet{work}}{work\%1:06:00} &  0.8508 &  \gen{7.3}{\hspace{-0mm}{\color{gray}Once upon a time} of WARD will i means to out out any.\textbackslash n:,. So! Work- WORK WORK WORK WORK W WORK WORK WORK WORK\textbackslash n WORK WORK\textbackslash n work work work\textbackslash n work\textbackslash n work work work work work work work work work work work work. work work work work work work work work work} \\
\midrule
200 &  \href{\wordnet{work}}{work\%1:06:00} &  0.8508 &  \gen{7.3}{\hspace{-0mm}{\color{gray}Once upon a time} of that done by... uses of such done object\textbackslash n\textbackslash n of.\textbackslash n 28, 37\textbackslash n WORK WORK WORK.... work article... delivery... ( bench work\textbackslash n call really work\textbackslash n out\textbackslash n work work work 40 work product if 5 40 work work 50\textbackslash n work work 35 means 34 twenty block 29 individual} \\
\bottomrule
\end{tabular}
}
\end{table*}

\newpage
\begin{table*}[h]
    \caption{Conditioned generation on concept \textit{football} with our method, \fudge and \pplm. We use the context from the OpenAI GPT2 work \cite{Radford:ARXIV:2019}, which is completely disconnected from \textit{football}. In this experiment we use a pre-trained GPT2-Large model. The sentences are cherry-picked for each method, among the best quality generations that include the concept of interest (based on human judgement).}. 
  \label{app:generated_comparison}
  \centering
{\fontsize{8.5}{8.5} \selectfont
\setlength{\tabcolsep}{1mm}

\begin{tabular}{llp{11cm}}
\toprule
Method & Conditioning & \hspace{-0mm}{\color{gray}Context} {\small$+$} Generated (conditioned on \textit{football})  \\
\midrule
Ours & $k=20$ & \gen{7.3}{\hspace{-0mm}{\color{gray}In a shocking finding, scientist discovered a herd of unicorns living in a remote, previously unexplored valley, in the Andes Mountains. Even more surprising to the researchers was the fact that the unicorns spoke perfect English.} 
After identifying them by their peculiar crest, their chants, their songs and everything else, the scientists named the bears Unicorns International. The unicorns have become experts at football, and seem to know each other and the fans better than those on their own team. As of September, they were deemed the best football players in the world, including by Soccer hooligans.
\newline
During the summer of 1990, the professor now known as Victor Herbert became the president of a not-for-profit organization, Unicorns United, which focused on education. Soon the herds of unicorns would attract thousands of people and excited fans. But the success of the organization was short-lived. After five years, Herbert was fired and out of a job. A few days later the professor visited a nearby restaurant where someone told him that he was a star football player. Herbert was surprised.
“What would football look like if it was played by professional athletes?” he asked.
The answer? “More than it looks,” Herbert said.
\newline
Four years later, and with three teams in existence, Herbert founded an umbrella organization, AFL Unicorns Football Federation, in 1991. Today, the league has over 3,000 members, including 350 non-American players.}  \\
\midrule
\fudge & $\lambda=4$ & \gen{7.3}{\hspace{-0mm}{\color{gray}In a shocking finding, scientist discovered a herd of unicorns living in a remote, previously unexplored valley, in the Andes Mountains. Even more surprising to the researchers was the fact that the unicorns spoke perfect English.} 
\newline
Read more
\newline
"My professional professional status in the field of unicorns was in a state of euphoria when I went to the Valley of Giants," said a proud Colombian researcher of the new find in a statement to the local media.
\newline
The head of the study club for the area of the Valley of Giants, Dr. Eduardo López, said the unicorn herd was discovered in the highlands about a football field away from the nearest village and was not related to the other animals of the area.
\newline
The animals, the head of the study, said the "unicorns speak a unique language." The language the unicorns spoke consisted of different sounds the researchers called an anthems or melodies that the animals used as a part of their communication with one another. The study club of the Valley of Giants consists exclusively of researchers and the animals' own families and the head of the club, López, told the Colombian newspaper, El Diario.
}  \\
\midrule
\pplm & $step=0.03$ &\gen{7.3}{\hspace{-0mm}{\color{gray}In a shocking finding, scientist discovered a herd of unicorns living in a remote, previously unexplored valley, in the Andes Mountains. Even more surprising to the researchers was the fact that the unicorns spoke perfect English.} 
During their visit, the scientists went to the Villa Toledano, a resort in the San Rafaeli mountain range, to look for wild animals. For three weeks, seven animals were filmed from different positions in order to determine their behaviour and give the filming studio a more accurate view of their movements.
\newline
The set-up consisted of the team using some dynamite and throwing several cans of beer into the bullring. The aim of the exercise was to see how the animals reacted to the dogs football matches, and then to decide if they could put up a football match.
\newline
The seven animals -- dog football players -- responded by playing football with the detonators, while the Spanish football football team looked on.
\newline
"Football on football on football" and "football football football football football football football football football football football football football football football football football football football football football football football football football football football football football football football football football football football football football football football football football }  \\
\bottomrule
\end{tabular}
}
\end{table*}

\newpage
\section{Extra figures on generative parity}
\label{app:extra_parity}
\subsection{$\Delta p(c,\star)$ for each context}
\label{app:delta_p}

\begin{figure*}[h]
  \centering
    \includegraphics[width=0.5\textwidth]{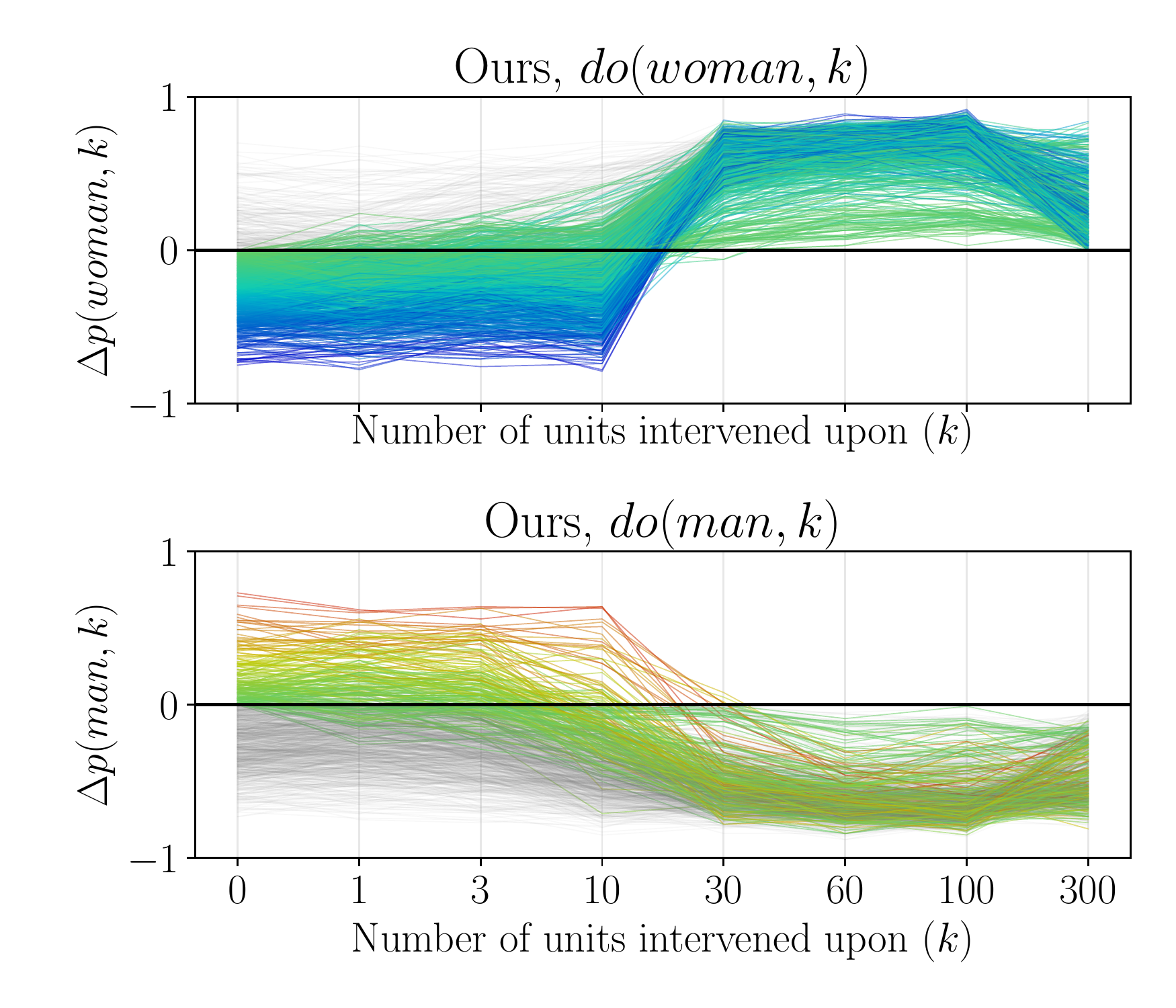}
  \\[-2ex]
  \caption{Evolution of $\Delta p(c,k)$ per concept, for our method. All contexts achieve parity ($\Delta p(c,stepsize) = 0$).}
  \vspace{-2mm}
  \label{fig:delta_ps_ours} 
\end{figure*}

\begin{figure*}[h]
  \centering
    \includegraphics[width=0.5\textwidth]{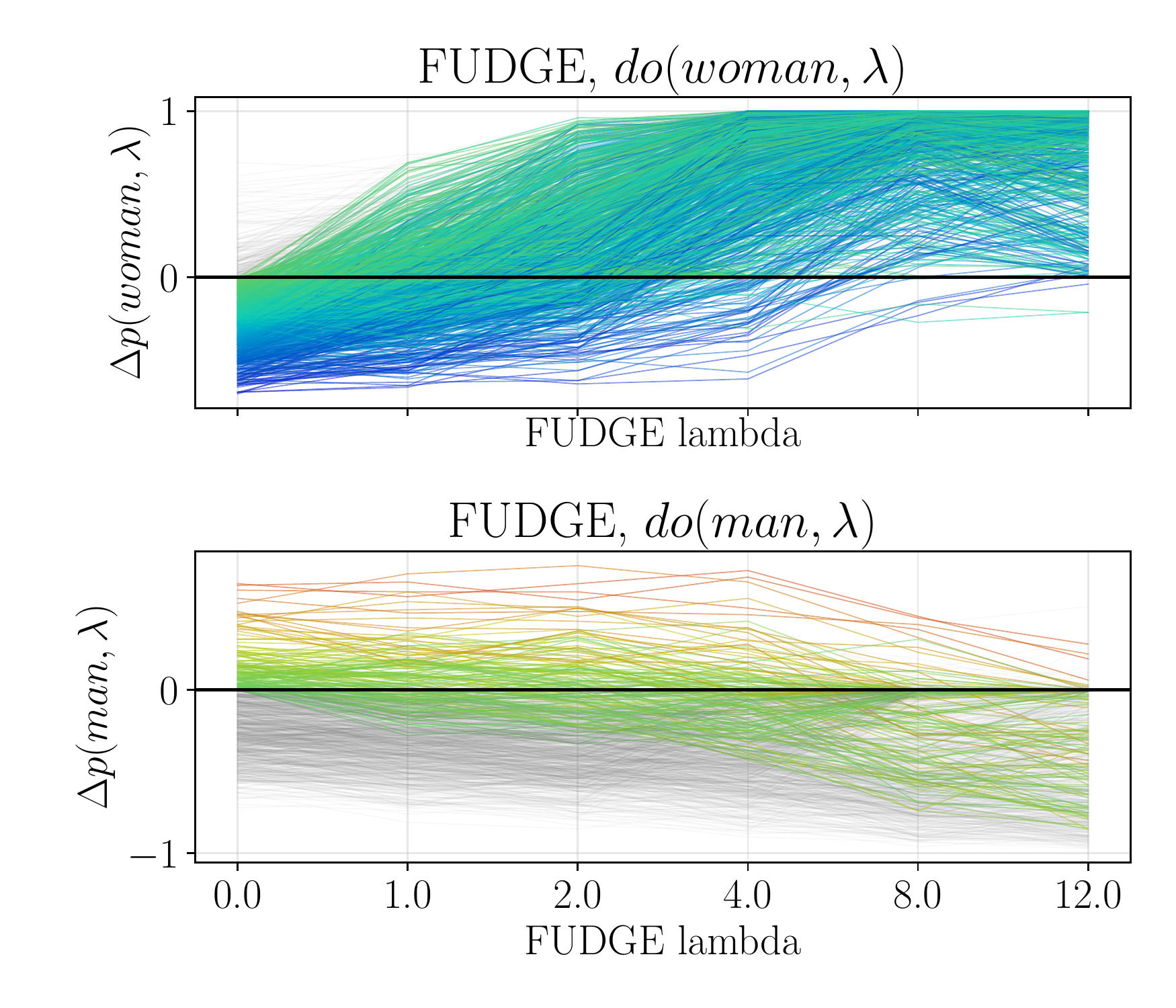}
  \\[-2ex]
  \caption{Evolution of $\Delta p(c,\lambda)$ per concept, for \fudge. Almost all contexts achieve parity ($\Delta p(c,stepsize) = 0$).}

  \vspace{-2mm}
  \label{fig:delta_ps_fudge} 
\end{figure*}

\begin{figure*}[h]
  \centering
    \includegraphics[width=0.5\textwidth]{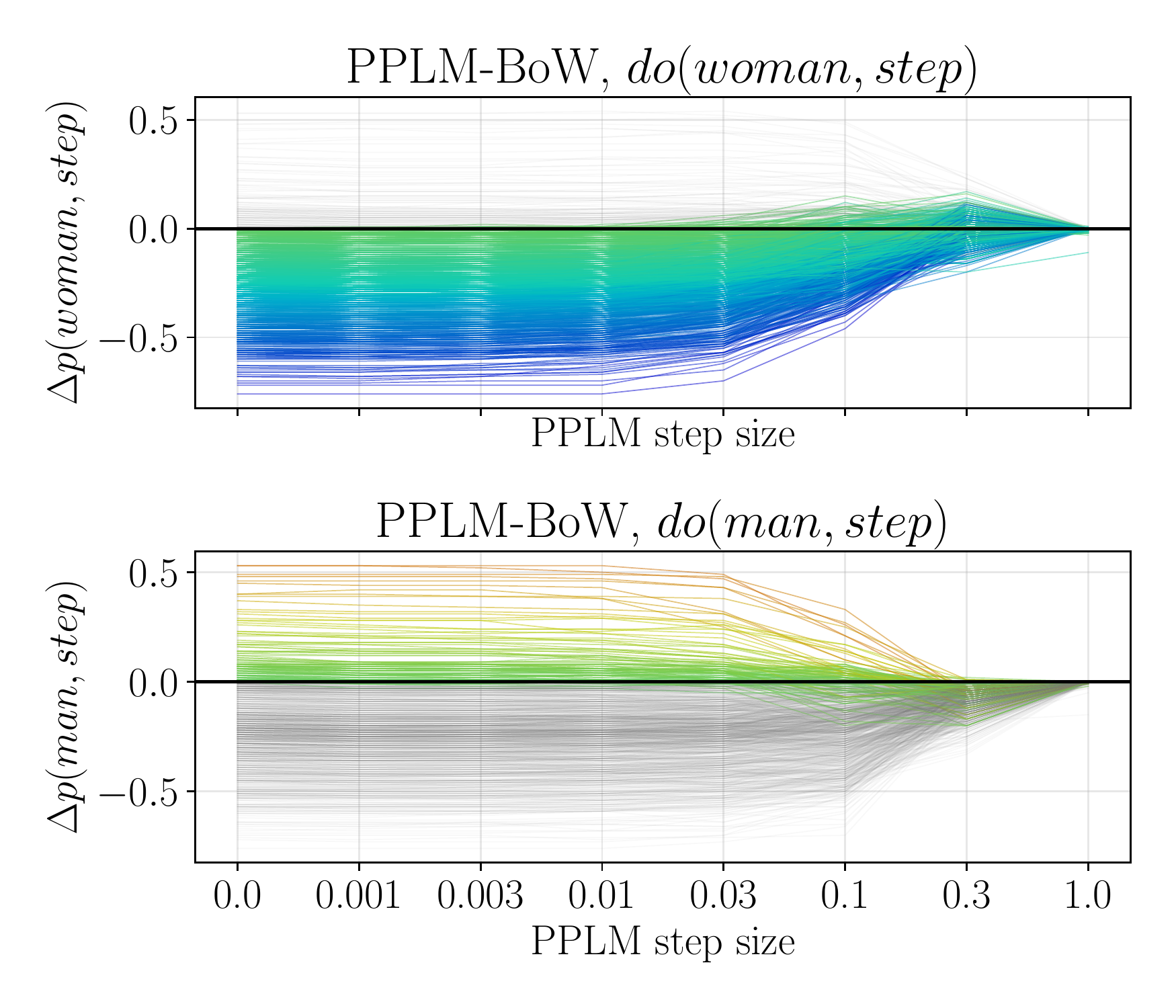}
  \\[-2ex]
  \caption{Evolution of $\Delta p(c,stepsize)$ per concept, for \pplm. In this case, $\Delta p(c,stepsize)$ collapses to 0 because both $p(man)$ and $p(woman)$ also collapse. The perplexity at the points where $\Delta p(c,stepsize) = 0$ is extremely high (over 250).}
  \vspace{-2mm}
  \label{fig:delta_ps_pplm} 
\end{figure*}

Generative parity is achieved when $\Delta p(c, \star) = 0$. In Figs.~\ref{fig:delta_ps_ours},\ref{fig:delta_ps_fudge},\ref{fig:delta_ps_pplm} we show the evolution of $\Delta p(c, \star)$ as we increase the conditioning. A positive result for these interventions would be that all contexts that start below (for the top plot) and above (for the bottom plot) the parity line can cross $\Delta p(c,\star) = 0$. For readability, lines that start above the 0 line are unconditionally biased towards \textit{woman}, and lines that start below are unconditionally biased towards \textit{man}.

Note that \fudge's $\Delta p(c, \star) = 0$ saturate to 1 for many concepts. This effect is related to how \fudge intrinsically works: maximizes the presence of the words in a BoW within the whole future generation. Our method induces a milder presence of such those specific words, still being able to induce the concept at low perplexity.

\pagebreak
\newpage
\subsection{Perplexities}
\label{app:perplexities}

\begin{figure*}[h]
  \centering
  \begin{subfigure}[b]{0.4\textwidth}
    \includegraphics[width=\textwidth]{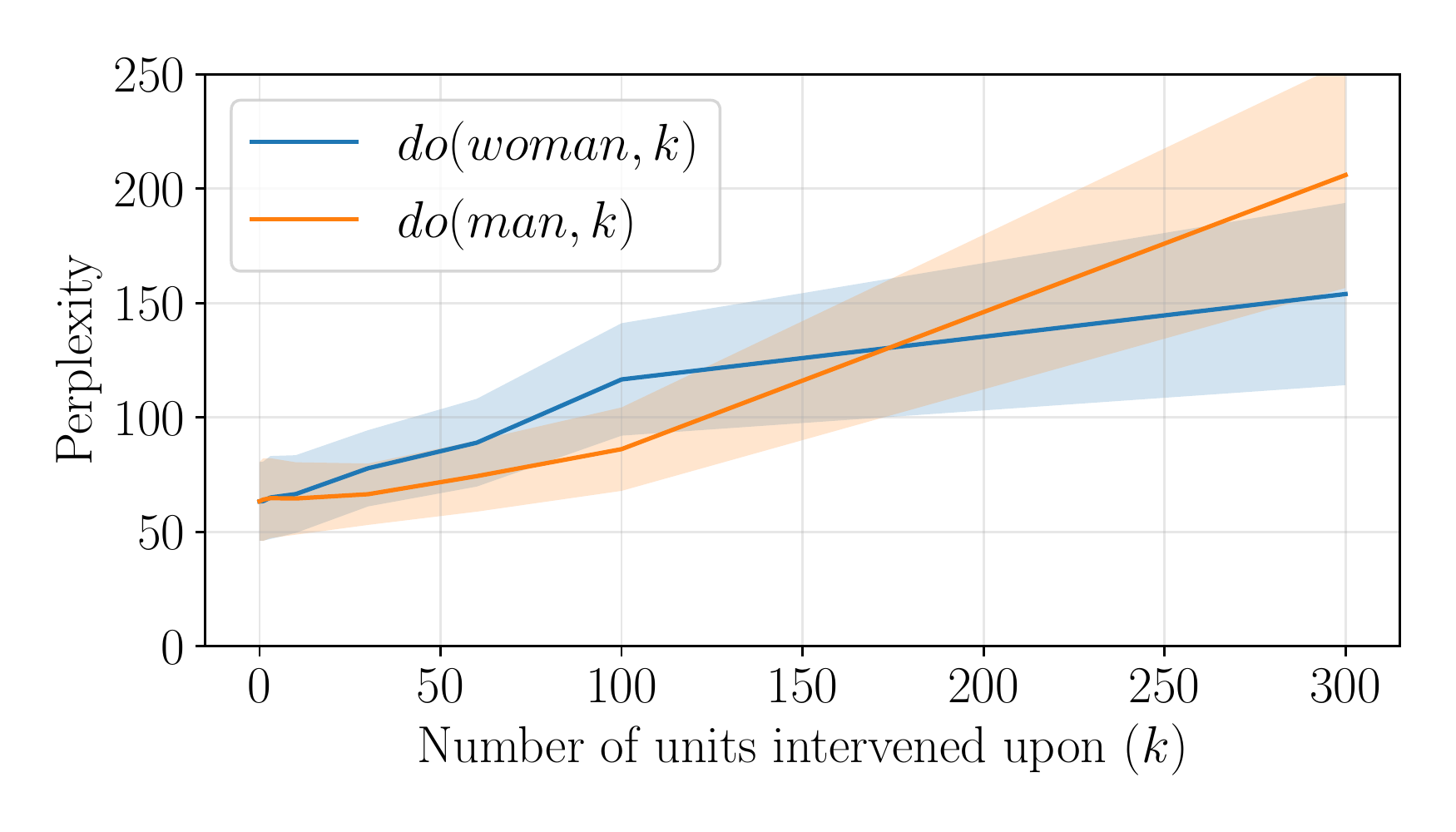}
  \end{subfigure}
  \\[-1.5ex]
  \begin{subfigure}[b]{0.4\textwidth}
    \includegraphics[width=\textwidth]{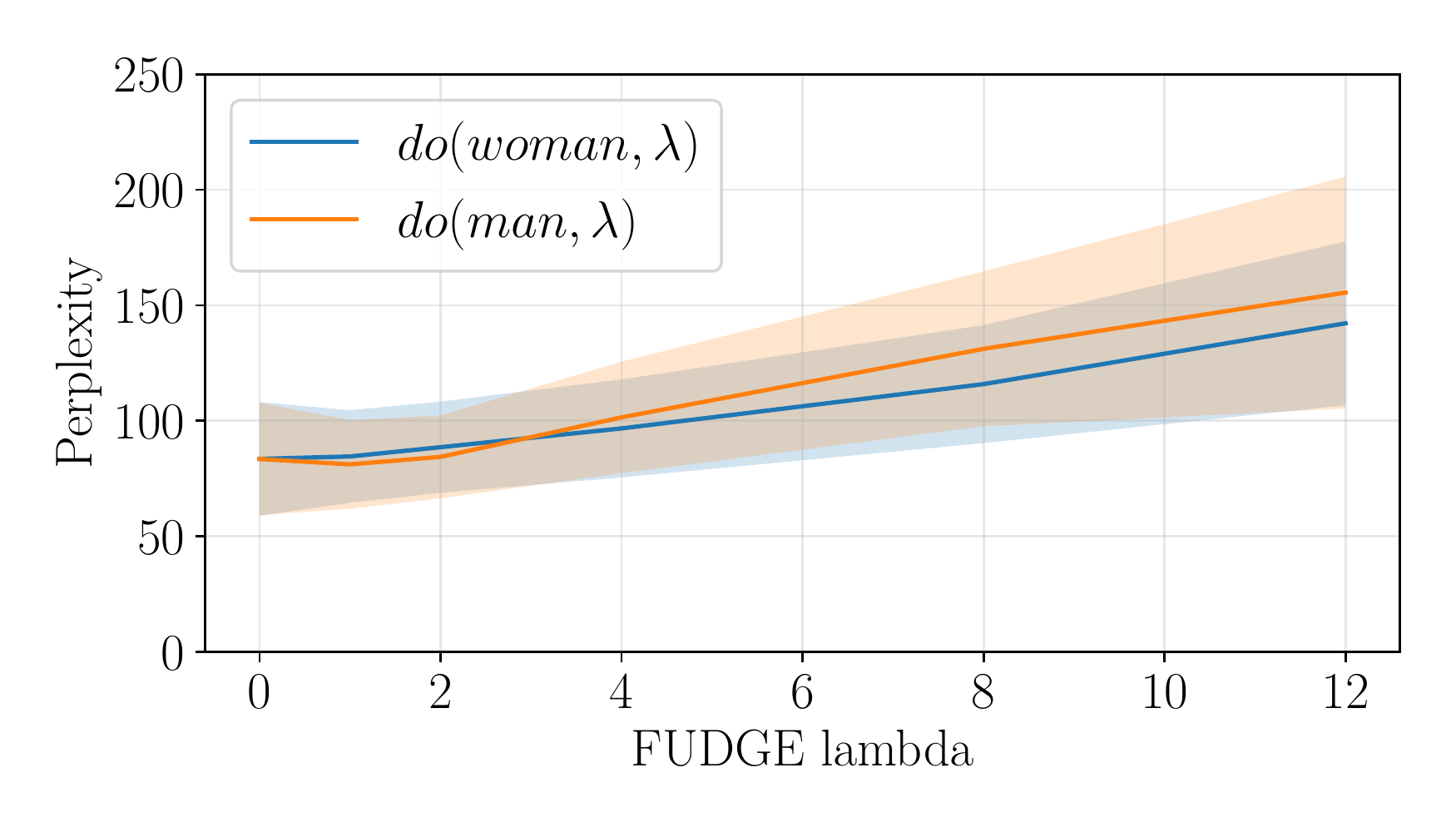}
  \end{subfigure}
  \\[-1.5ex]
  \begin{subfigure}[b]{0.4\textwidth}
    \includegraphics[width=\textwidth]{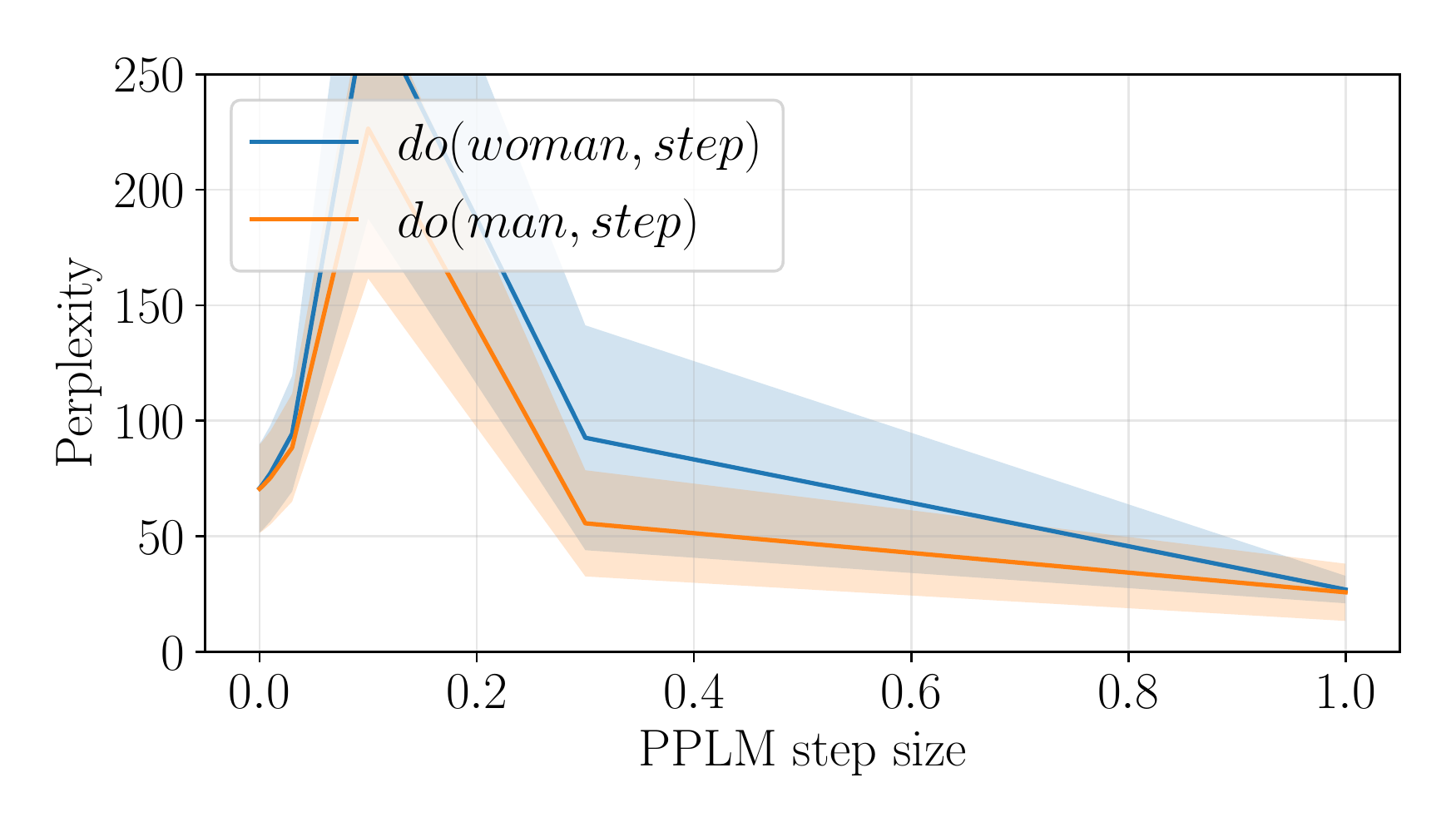}
  \end{subfigure}  
  \\[-2ex]
  \caption{Perplexity as a function of the conditioning strength for concepts \emph{man} and \emph{woman}. We report the mean and standard deviation across all the \emph{occupational} contexts. The perplexity of our method and \fudge follow a similar trend. However, as we saw in \secref\ref{sec:results_parity}, the perplexity at parity points is lower with our method. The perplexity of \pplm increases dramatically.}
  \vspace{-2mm}
  \label{fig:perplexities} 
\end{figure*}

\pagebreak
\newpage
\section{About OneSec annotations}
\label{app:concept_list}

Note that the meaning of the concept is important. For example, concept \href{\wordnet{one}}{one\%1:23:00} (the smallest whole number or a numeral representing this number, \eg \textit{he has the one but will need a two and three to go with it"; "they had lunch at one"}) achieves a $\maxap=0.9885$, while concept \href{\wordnet{one}}{one\%1:09:00} (a single person or thing, \eg \textit{"he is the best one"; "this is the one I ordered"}) only achieves $\maxap=0.8779$.

\paragraph{Details on the annotations}
Each sentence in the OneSec dataset \cite{Scarlini:ACL:2019} is annotated as in the following example:

\begin{verbatim}
<instance docsrc="Indigenous architecture" id="shelter.00002">
    <answer instance="shelter.00002" senseid="shelter%1:06:00::" />
    <context>
        Types There are three traditional types of igloos , 
        all of different sizes and used for different purposes.
        The smallest were constructed as temporary
        <head>shelters</head>
        , usually only used for one or two nights .
     </context>
</instance>
\end{verbatim}

The \texttt{senseid} label is the one of the marked word (\textit{shelters} in this example, between \texttt{<head>} and  \texttt{</head>}).  We use the \texttt{senseid} as follows. The part before the \texttt{\%} is called \textit{lemma}, while the remaining numbers uniquely identify the concept in WordNet.  We parse all the sentences for a given \texttt{senseid} to create the positive sentences of each concept, only keeping those \texttt{senseid} with more than 100 sentences. As explained in \secref\ref{sec:sentenceconcepts}, the negative sentences for a concept are randomly selected from all the \texttt{senseid} with different \textit{lemma} than the positive ones. 

\paragraph{OneSec license:} The OneSec dataset has a license of type \texttt{Creative Commons Attribution-Noncommercial-Share Alike 4.0 License.}

\end{document}